\def\isarxiv{1} 

\ifdefined\isarxiv
\documentclass[11pt]{article}

\usepackage[numbers]{natbib}

\else
\documentclass{article}
\usepackage{microtype}
\usepackage{graphicx}
\usepackage{subfigure} 
\usepackage{hyperref}
\usepackage{icml2024} 
\fi

\usepackage{amsmath}
\usepackage{amsthm}
\usepackage{amssymb}
\usepackage{algorithm}
\usepackage{algpseudocode}
\usepackage{grffile}
\usepackage{wrapfig,epsfig}
\usepackage{url}
\usepackage{xcolor}
\usepackage{epstopdf}

\usepackage{bbm}
\usepackage{dsfont}

\allowdisplaybreaks

\ifdefined\isarxiv

\usepackage{tikz}
\usepackage{hyperref}  
\hypersetup{colorlinks=true,citecolor=blue,linkcolor=blue} 
\usetikzlibrary{arrows}
\usepackage[margin=1in]{geometry}

\else

\fi

\newtheorem{theorem}{Theorem}[section]
\newtheorem{lemma}[theorem]{Lemma}
\newtheorem{definition}[theorem]{Definition}

\newtheorem{corollary}[theorem]{Corollary}

\newtheorem{fact}[theorem]{Fact}
\newtheorem{remark}[theorem]{Remark}

\newcommand{\wt}{\widetilde}
\newcommand{\ov}{\overline}
\newcommand{\N}{\mathcal{N}}
\newcommand{\R}{\mathbb{R}}

\newcommand{\sketch}{\mathrm{sketch}}
\renewcommand{\d}{\mathrm{d}}

\renewcommand{\tilde}{\wt}

\newcommand{\ReLU}{\mathsf{ReLU}}
\newcommand{\NTK}{\mathsf{NTK}}

\DeclareMathOperator*{\E}{{\mathbb{E}}}

\DeclareMathOperator{\poly}{poly}

\DeclareMathOperator{\dis}{dis}
\DeclareMathOperator{\cts}{cts}

\makeatletter
\newcommand*{\RN}[1]{\expandafter\@slowromancap\romannumeral #1@}
\makeatother

\usepackage{lineno}

\begin{document}

\ifdefined\isarxiv

\title{Training Overparametrized Neural Networks in Sublinear Time}
\date{}

\author{
Yichuan Deng\thanks{\texttt{ycdeng@cs.washington.edu}. The University of Washington.}.
\and 
Hang Hu\thanks{\texttt{h-hu19@mails.tsinghua.edu.cn}. Tsinghua University.}
\and 
Zhao Song\thanks{\texttt{zsong@adobe.com}. Adobe Research.}
\and 
Omri Weinstein\thanks{\texttt{omri@cs.columbia.edu}. The Hebrew University and Columbia University.}
\and 
Danyang Zhuo\thanks{\texttt{danyang@cs.duke.edu}. Duke University.}
}

\else

\icmltitlerunning{Training Overparametrized Neural Networks in Sublinear Time}

\twocolumn[

\icmltitle{Training Overparametrized Neural Networks in Sublinear Time}


\icmlsetsymbol{equal}{*}

\begin{icmlauthorlist}
\icmlauthor{Aeiau Zzzz}{equal,to}
\icmlauthor{Bauiu C.~Yyyy}{equal,to,goo}
\icmlauthor{Cieua Vvvvv}{goo}
\icmlauthor{Iaesut Saoeu}{ed}
\icmlauthor{Fiuea Rrrr}{to}
\icmlauthor{Tateu H.~Yasehe}{ed,to,goo}
\icmlauthor{Aaoeu Iasoh}{goo}
\icmlauthor{Buiui Eueu}{ed}
\icmlauthor{Aeuia Zzzz}{ed}
\icmlauthor{Bieea C.~Yyyy}{to,goo}
\icmlauthor{Teoau Xxxx}{ed}\label{eq:335_2}
\icmlauthor{Eee Pppp}{ed}
\end{icmlauthorlist}

\icmlaffiliation{to}{Department of Computation, University of Torontoland, Torontoland, Canada}
\icmlaffiliation{goo}{Googol ShallowMind, New London, Michigan, USA}
\icmlaffiliation{ed}{School of Computation, University of Edenborrow, Edenborrow, United Kingdom}

\icmlcorrespondingauthor{Cieua Vvvvv}{c.vvvvv@googol.com}
\icmlcorrespondingauthor{Eee Pppp}{ep@eden.co.uk}

\icmlkeywords{Machine Learning, ICML}

\vskip 0.3in
]

\printAffiliationsAndNotice{\icmlEqualContribution} 


\fi

\ifdefined\isarxiv
\begin{titlepage}
  \maketitle
  \begin{abstract}

The success of deep learning comes at a tremendous computational and energy cost, and the scalability of training massively overparametrized neural networks is becoming a real  barrier to the progress of artificial intelligence (AI). Despite the popularity and low cost-per-iteration of traditional backpropagation via gradient decent, stochastic gradient descent (SGD) has prohibitive convergence rate in non-convex settings, both in theory and practice.  

To mitigate this cost, recent works have  proposed to employ alternative (Newton-type) training methods with much faster convergence rate, albeit with higher cost-per-iteration. 
For a typical neural network with $m=\mathrm{poly}(n)$ parameters and input batch of  $n$  datapoints in $\mathbb{R}^d$, the previous work of \cite{bpsw21} requires $\sim mnd + n^3$ time per iteration. In this paper, we present a novel training method that requires only $m^{1-\alpha} n d + n^3$ amortized time in the same overparametrized regime, where $\alpha \in (0.01,1)$ is some fixed constant. This method relies on a new and alternative view of neural networks, as a set of binary search trees, where each  iteration corresponds to modifying a small subset of the nodes in the tree. We believe this view would have further applications in the design and analysis of deep neural networks (DNNs). 

  \end{abstract}
  \thispagestyle{empty}
\end{titlepage}

{\hypersetup{linkcolor=black}
\tableofcontents
}
\newpage

\else



\begin{abstract}

The success of deep learning comes at a tremendous computational and energy cost, and the scalability of training massively overparametrized neural networks is becoming a real  barrier to the progress of artificial intelligence (AI). Despite the popularity and low cost-per-iteration of traditional backpropagation via gradient decent, stochastic gradient descent (SGD) has prohibitive convergence rate in non-convex settings, both in theory and practice.  

To mitigate this cost, recent works have  proposed to employ alternative (Newton-type) training methods with much faster convergence rate, albeit with higher cost-per-iteration. 
For a typical neural network with $m=\mathrm{poly}(n)$ parameters and input batch of  $n$  datapoints in $\mathbb{R}^d$, the previous work of \cite{bpsw21} requires $\sim mnd + n^3$ time per iteration. In this paper, we present a novel training method that requires only $m^{1-\alpha} n d + n^3$ amortized time in the same overparametrized regime, where $\alpha \in (0.01,1)$ is some fixed constant. This method relies on a new and alternative view of neural networks, as a set of binary search trees, where each  iteration corresponds to modifying a small subset of the nodes in the tree. We believe this view would have further applications in the design and analysis of deep neural networks (DNNs). 
\end{abstract}

\fi

\section{Introduction}
Deep learning technology achieves unprecedented accuracy across many domains of AI and human-related tasks, from computer vision, natural language processing, and robotics. 
This success, however, is approaching its limit and is largely compromised by the computational complexity of these resource-hungry models.  
State-of-art neural networks keep growing larger in size, requiring giant matrix operations to train billions of parameters \cite{dclt18,rwc+19,bmr+20,cnd+22,zrg+22,cha22,o23}.  This barrier is  exacerbated by the empirical phenomenon that  \emph{overparametrization} in DNNs \cite{jgh18} keeps improving model accuracy, despite the danger of overfitting \cite{double_descent}, motivating the design of complex networks which need to train billions of parameters. 
As such, scalable training of deep neural networks is a major challenges of modern AI \cite{wrg+22,ss17}.

Training a neural network can be broadly viewed a greedy iterative process,  starting from an initial set of weight matrices (one per layer of the network). In each iteration, the algorithm chooses a (possibly complicated) rule for updating   the value of current weights $W_i$ based on the training data, yielding the new weight matrices $W_{i+1}$. The total running time of DNN training is generally composed of two parts: The \emph{number of iterations} (i.e., convergence rate) and the  \emph{cost-per-iteration} (i.e., $\mathsf{CPI}$). A long line of research in convex and non-convex optimization has focused on the former question \cite{k80,k84,v87,r88,v89,m13_flow,ls14,m16,lsz19,jlsw20,hjstz22,dsx23,dsy24}. This paper's focus is on the latter question. 

The most popular iterative method for training DNNs is via  \emph{stochastic gradient descent} and its regularized variations \cite{ll18,dzps19,als19a,als19b,sy19,wdw19}.
The popularity of this method is justified, to a great extent, by the simplicity and fast $\mathsf{CPI}$. 
Calculating the gradient of the loss function is linear in the dimension of the gradient in each iteration, especially with mini-batch sampling \cite{hrs16,cgh+19}). Alas, the theoretical convergence rate (number of iterations) of first-order methods  
is dauntingly slow in \emph{non-convex} landscapes due to  pathological curvatures ($\Omega(\poly(n)\log(1/\epsilon))$ for reducing the training error below $\epsilon$ in overparametrized networks, see e.g.,~\cite{zmg19}). 

A recent line of work proposed to mitigate this drawback by replacing (S)GD with \emph{second-order} (Newton-type) methods, which exploit information of the Hessian (curvature) of the loss function,  
and are proven to converge dramatically faster, at a rate of $O(\log (1/\epsilon) )$ iterations, which is \emph{independent} of the input size \cite{mg15,zmg19}.
In contrast, Newton methods have a high $\mathsf{CPI}$. 
since they need to compute the \emph{inverse} of Hessian matrix, which is dense and changes dynamically.
The recent works of \cite{cgh+19, zmg19} showed that this computational bottleneck can be mitigated for overparametrized DNNs ($m=\poly(n)$) with smooth (resp. $\ReLU$) activations, and presented a Gauss-Newton (resp. NGD) training algorithm with $O(mn^2)$ training time per iteration. Here $m$ is the number of neurons. We let $n$ be the number of inputs. This runtime was further improved in the work of \cite{bpsw21}, who showed how to implement the Gauss-Newton algorithm in $O(mnd + n^3)$ time per iteration, which is \emph{linear time} in the network size, assuming $m \gtrsim n^2$ (as the dimensions of the Jacobian matrix of the loss is $\Theta(mnd)$ without simplifying assumptions \cite{mg15}).

It is tempting to believe that linear-time per iteration \cite{bpsw21} is unavoidable --  For a network with $m$ neurons and a training set of $n$  points in $\R^d$, each iteration spends at least $\sim nmd$ time to go through each training datapoint and each neuron. Indeed, this was a common feature of all aforementioned training methods.

Nevertheless, in this paper we present a novel training method with \emph{sublinear} cost per iteration in the network size, while retaining \emph{the same convergence rate} (number of iterations) as the prior state-of-art methods \cite{bpsw21, zmg19, cgh+19}.
More formally, let $f: \R^d \rightarrow \R$ be a neural network  
\[
    f(x) := \sum_{r=1}^m a_r \cdot \phi ( \langle w_r, x \rangle - b )
\] 
with bias $b > 0$, $a \in \{ \pm 1\}^m$, each $w_r \in \R^d$, for all $r\in [m]$. Our main resuilt is as follows. 

\begin{theorem}[Main Result, Informal] \label{thm:main}
Suppose there are $n$ training data points in $\R^d$.
Let $f_{m,n}$ be a sufficiently wide two-layer $\ReLU$ $\mathsf{NN}$ with $m = \poly(n)$ neurons. Let $\alpha \in (0.01,1)$ be some fixed constant.
Let $\epsilon \in (0,0.1)$ be an accuracy parameter.  Let ${\cal T}(\epsilon)$ denote the overall time for shrinking loss down to $\epsilon$. 
There is a (randomized) algorithm (Algorithm~\ref{alg:ours_informal})  that, with probability $1-1/\poly(n)$, reduces the training error by $1/2$ in each iteration (note that $f_t$ is $f_{m,n}$ at time $t$) 
\begin{align*}
    \ell_2\mathrm{-loss}( f_{t+1}, y ) \leq \frac{1}{2} \cdot \ell_2\mathrm{-loss}( f_t , y )
\end{align*}
in amortized cost-per-iteration ($\mathsf{CPI}$)
\[ \wt{O}( m^{1-\alpha} nd +  n^3).\] 

The overall running time (including initialization) ${\cal T}(\epsilon)$ is 
\begin{align*}
    O(mnd) + \wt{O}(  m^{1-\alpha}nd +  n^3) ) \cdot \log (1/\epsilon).
\end{align*}

If the algorithm is allowed to use fast matrix multiplication ($\mathsf{FMM}$), then
the $\mathsf{CPI}$ becomes 
\begin{align*}
    \wt{O} ( m^{1-\alpha}nd + n^\omega),
\end{align*}
and the ${\cal T}(\epsilon)$ becomes 
\begin{align*}
    O(mnd) +  \wt{O}( m^{1-\alpha} nd +    n^\omega) \cdot \log (1/\epsilon),
\end{align*}
where $\omega$ is the exponent of matrix multiplication, which is currently approximately equal to $2.373$.

The randomness is from two parts: the first part is random initialization weights, and the second part is due to internal randomness of our algorithm.
\end{theorem}

\begin{remark}
Notice that the linear cost term $O(mnd)$ for merely computing the network's loss matrix, is only incurred once at the \emph{initialization} of our training algorithm, whereas in  \cite{bpsw21} and all prior work \cite{cgh+19, zmg19}, this linear term is payed \emph{every iteration} (i.e., $T\cdot mnd$ as opposed to our $T + mnd$). 
Our theorem therefore provides a direct improvement over \cite{bpsw21} when $m = \poly(n)$.
\end{remark}

\paragraph{Key Insight: DNNs as Binary-Search Trees}
Our algorithm is based on an alternative view of DNNs, as a \emph{set of binary search trees}, where the relationship between the network's weights and a training data point is encoded using a binary tree: Each leaf represents the inner product of a neuron and the training data, and each intermediate (non-leaf) node represents the \emph{larger} out of the left and right child. This simple yet new representation of neural networks turns out to enable fast training -- The centerpiece of our result is an analysis proving that in each iteration, only a small subset $K$ of paths in this tree collection needs to be updated (amortized worst-case), due to the \emph{sparsity} of activations. Consequently, we only need to update $nK \log m$ tree nodes per iteration. In the Technical Overview Section \ref{sec:tech_ow}, we elaborate more on its details.

\paragraph{Roadmap.} We describe the organization of this work in next a few sentences. We propose our main problem and present the tools we need to use in Section \ref{sec:preli}. In Section \ref{sec:tech_ow}, we specifically overview the techniques used in this paper. In Section \ref{sec:correctness}, we analyze the correctness of our algorithm, specifically, we prove the training loss converges. In Section \ref{sec:running_time}, we analyze the running time of our algorithm.  In Section \ref{sec:conclusion}, we provide a conclusion. 

\section{Preliminaries} \label{sec:preli}

\subsection{Notations.}
For any positive integer $n$, we use $[n]$ to denote set $\{1,2,\cdots,n\}$. For any function $f$, we use $\wt{O}(f)$ to denote $f \cdot \poly (\log f)$. For two vectors $w$ and $x$, we use $\langle w, x \rangle$ to denote inner product. We use $a^\top$ to denote the transpose of $a$. We use $\E[]$ to denote expectation and $\Pr[]$ for probability. For convenience, we use $\mathsf{FMM}$ to denote fast matrix multiplication. We use $\NTK$ to denote neural tangent kernel. We use $\ReLU$ to denote rectified linear unit. We use $\mathsf{NN}$ to denote neural network. We use $\mathsf{CPI}$ to represent the cost per iteration. We use PSD to denote positive semidefinite.

\subsection{A Sketching Tool
}
\label{sec:app_prel:sketch}
Sarl\'{o}s \cite{s06} firstly introduced the notation of subspace embedding. Many numerical linear algebra applications have used that concept and its variations \cite{cw13,nn13,rsw16,swz17,swz19b,sy21}.
The formal definition is:
\begin{definition}[Oblivious subspace embedding, {\sf OSE} \cite{s06}]\label{def:ase1}
	Given an $N \times k$ matrix $B$, an $(1 \pm \epsilon)$ $\ell_2$-subspace embedding for the column space of $B$ is a matrix $S$, such that for any $x \in \R^k$, 
	\begin{align*} 
	(1 - \epsilon) \| B x \|_2^2 \le \| S B x \|_2^2 \le (1 + \epsilon) \| B x \|_2^2.
	\end{align*}
	Equivalently, let $U$ be the matrix whose columns form an orthonormal basis containing the column vectors of $B$, then
	\begin{align*} 
	\| I - U^\top S^\top S U \|_2 \leq \epsilon.
	\end{align*}
\end{definition}

It is known that subspace embedding can be given by a Fast-JL sketching matrix \cite{ac06,dmm06,t11,dmmw12,ldfu13,psw17} with a classical $\epsilon$-net argument, 

\begin{lemma}
    \label{lem:subspace-embedding}
    Assume that $N = \poly(k)$. Assume $\delta \in (0,0.1)$. 
    For a matrix $B \in \R^{N \times k}$, we can produce an $(1\pm \epsilon)$ $\ell_2$-subspace embedding $S \in \R^{k \poly ( \log (k/\delta) )/\epsilon^2 \times N}$ for $B$ with probability at least $1-\delta$.
	
	In addition, $SB$ takes $O(Nk\cdot \poly\log k)$ time to be generated.
\end{lemma}

\subsection{Model formalization} \label{sec:model_formalization}
In this section, we formalize the $\mathsf{NN}$ model and the main problem of this paper. 
When there is no ambiguity, we will always use the notations in this section throughout the whole paper.

We first define the 2-layer $\ReLU$ activated neural network and its loss function.

\begin{definition}[2-layer $\ReLU$ activated neural network] \label{def:neural_network}
    Suppose the dimension of input is $d$, the number of intermediate nodes (or hidden neurons) is $m$, the dimension of output is $1$, the batch size is $n$ and the shifted parameter is $b$ ($b \ge 0$). Then the weight of the first layer can be characterized by $m$ $d$-dimensional vectors $w_1,w_2,\cdots,w_m$, and the weight of the second layer can be characterized by $m$ scalars $a_1,a_2,\cdots,a_m$. For convenience, define 
    \begin{align*}
        W = [~w_1^\top~w_2^\top~\cdots~w_m^\top~]^\top
    \end{align*}
    and
    \begin{align*}
        a = [~a_1~a_2~\cdots~a_m~]^\top,
    \end{align*}
    given an input $x\in\R^d$, the 2-layer $\ReLU$ activated neural network outputs
    \begin{align*}
        f(W,x,a) = \frac{1}{\sqrt{m}} \sum_{r=1}^m a_r \phi(\langle w_r,x \rangle)
    \end{align*}
    where $\phi(x) = \max\{x,b\}$ is called shifted $\ReLU$ activation function.

    For simplicity, we suppose the data is normalized, that is, $\|x\|_2 = 1$. This is natural in both practical machine learning, and machine learning theory.
    
    We also suppose $a \in \{-1,+1\}^m$ is fixed throughout training. This is also natural in the area of theoretical deep learning \cite{ll18,dzps19,als19a,all19,sy19,bpsw21,z22}.
\end{definition}

\begin{definition}[Loss function]
    Suppose the dimension of input is $d$, the number of intermediate nodes (or hidden neurons) is $m$, the dimension of output is $1$, the batch size is $n$ and the shifted parameter is $b$ ($b \ge 0$). 
    For a fixed set of $n$ 
    points $x_1 , x_2, \cdots, x_n \in \R^{d}$ and their corresponding labels $y_1, y_2, \cdots, y_n \in \R$. Consider the following loss function:
    \begin{align*}
        \mathcal{L} (W) := \frac{1}{2} \sum_{i=1}^n ( y_i - f (W,x_i,a) )^2 .
    \end{align*}
\end{definition}

By mathematics, 
\begin{align}\label{eq:relu_derivative}
\frac{\partial f (W,x,a)}{\partial w_r}=\frac{1}{ \sqrt{m} } a_r x{\bf 1}_{ w_r^\top x \geq b }, ~~~ \forall r \in [m].
\end{align}

Thus one can calculate the gradient of loss function $\mathcal{L}$
\begin{align}\label{eq:gradient}
         \frac{ \partial \mathcal{L}(W) }{ \partial w_r } 
    =  \frac{1}{ \sqrt{m} } \sum_{i=1}^n ( f(W,x_i,a) - y_i ) \cdot a_r \cdot x_i \cdot {\bf 1}_{ w_r^\top x_i \geq b }.
\end{align}

Then we define the prediction function, the Jacobi matrix, and the Gram matrix.
\begin{definition}[prediction function]
    For a batch of inputs $\{ (x_i , y_i) \}_{i\in [n]} \in \R^{d} \times \R$, we denote 
    \begin{align*}
        \alpha_{r, i}(t) := \phi(\langle w_r(t), x_i \rangle)
    \end{align*}
    for every $r \in [m]$ and $i \in [n]$.
    
    Prediction function $f_t : \R^{d \times n} \rightarrow \R^n $ at time $t$ is defined as follows 
    \begin{align*}
        f_t  = \frac{1}{\sqrt{m}} \sum_{r \in [m]}
        \begin{bmatrix}
          a_r \cdot \alpha_{r, 1}(t) ) \\
           a_r \cdot \alpha_{r, 2}(t) ) \\
        \vdots \\
           a_r \cdot \alpha_{r, n}(t) ) \\
        \end{bmatrix}
    \end{align*}
    Note that $w_r(t)$ is the $r$-th weight of the first layer after training of $t$ times. 
    
    For convenience, we define weight matrix \begin{align*} 
    W_t = [ w_1(t) ~ w_2(t) ~ \cdots ~ w_m(t) ] \in \R^{d \times m}
    \end{align*}
    In addition, we write 
    data matrix
    \begin{align*} 
        X = [x_1 ~ x_2 ~ \cdots ~ x_n] \in \R^{d \times n}.
    \end{align*} 
\end{definition}

\begin{definition}[Jacobi matrix and related definitions]
For each $i \in [n]$, $r \in [m]$ and $t \in [T]$, we define
\begin{align*}
    \beta_{r,i}(t) := {\bf 1}_{ \langle w_r(t), x_i \rangle \geq b }.
\end{align*}
    For every time step $t$, we use $J_t \in \R^{n \times m}$ to denote the Jacobian matrix at $t$. Formally, it can be written as
    {\small
    \begin{align*}
        &~J_t \\
        = &~\frac{1}{ \sqrt{m} }
        \left[
        \begin{matrix}
            a_1 x_1^{\top} \beta_{1,1}(t) & a_2 x_1^{\top} \beta_{2,1}(t) &  \cdots  & a_m x_1^{\top}  \beta_{m,1}(t)  \\
            a_1 x_2^{\top} \beta_{1,2}(t) & a_2 x_2^{\top}  \beta_{2,2}(t) &  \cdots  & a_m x_2^{\top}  \beta_{m,2}(t)  \\
            \vdots & \vdots & \ddots & \vdots \\
            a_1 x_n^{\top} \beta_{1,n}(t) & a_2 x_n^{\top} \beta_{2,n}(t) & \ldots & a_m x_n^{\top} \beta_{m,n}(t) \\
        \end{matrix}
        \right].
    \end{align*}
    }
    For each $i\in[n]$, we define $J_t(x_i)$ as the $i$-th row of $J_t$.
\end{definition}

\begin{definition}[Gram matrix]
Let $G_t \in \R^{n \times n}$ denote the Gram matrix. 
    Then $G_t$ can be formally written as $G_t = J_t J_t^{\top}$. The $(i, j)$-th entry of $G_t$ is the inner product between gradient in terms of $x_i$ and the gradient in terms of $x_j$, i.e.,
    \begin{align*} 
    (G_t)_{i,j} := \langle \frac{f(W_t, x_i)}{\partial W}, \frac{f(W_t, x_j)}{\partial W} \rangle.
    \end{align*}
\end{definition}
\cite{jgh18, dzps19, syz21} gave a crucial observation that the asymptotic of the Gram matrix $G$ is equal to a PSD matrix $K \in \R^{n \times n}$. The formal definition is 
\begin{align}
\label{eq:kernel}
K(x_i, x_j) := \E_{w\sim \mathcal{N}(0, I)}\left[x_{i}^{\top}x_{j}\textbf{1}_{\langle w, x_i \rangle \geq b, \langle w, x_j \rangle \geq b} \right].
\end{align}
\cite{jgh18,dzps19} only consider the case where $b=0$ and \cite{syz21} consider the general case $b \geq 0$.

\begin{remark}
    We use $\lambda$ to denote the minimal eigenvalue of the kernel matrix $K$ defined in Eq.~\eqref{eq:kernel}.
\end{remark}

\subsection{Problem definition}
We formalize our main problem as follows.
\begin{definition}[Main problem]
    The goal of this paper is to propose a training algorithm such that for an arbitrary 2-layer $\ReLU$ activated neural network defined in Definition \ref{def:neural_network}, it converges with high probability, and the running time of each iteration is sublinear in $nmd$ (i.e. $o(nmd)$).
\end{definition}

\subsection{Neural tangent kernel and its relation with data separability}
Neural Tangent Kernel ($\NTK$) is a Kernel matrix related to a multi-layer $\ReLU$ activated neural network. It is crucial in the analysis of Jacobi matrix. \cite{syz21} expanded the related concepts and revealed their properties, especially its relation to the data separability of an input batch.

As for data separability, it is a common assumption to the input of a neural network, and it has been used in many over-parameterized neural network literature \cite{ll18,als19a}. We first define kernels,
\begin{definition}
    Let $b \geq 0$ be the shift parameter. We define continuous version of the shifted $\NTK$ $H^{\cts}$ and discrete version of shifted $\NTK$ $H^{\dis}$ as 
    \begin{align*}
        H_{i,j}^{\cts} := & ~ \E_{ w \sim {\cal N}(0,I) } [ x_i^\top x_j {\bf 1}_{ w^\top x_i \geq b, w^\top x_j \geq b } ] , \\
        H_{i,j}^{\dis} := & ~ \frac{1}{m} \sum_{r=1}^m [ x_i^\top x_j {\bf 1}_{ w_r^\top x_i \geq b, w_r^\top x_j \geq b } ]. 
    \end{align*}
\end{definition}

Next, we define data separability,
\begin{definition}[Separability of input data] \label{def:separability}
Suppose we are given $n$ (normalized) input data points 
\begin{align*}
    \{x_1,x_2, \cdots, x_n \} \subseteq \R^d.
\end{align*}
Assume those points satisfy that $\forall i \in [n], \| x_i\|_2=1$. For each $i,j$, we define 
\begin{align*}
    \delta_{i,j}^+ = x_i + x_j\text{~and~}\delta_{i,j}^- =x_i -x_j. 
\end{align*}
Let $\delta$ be the data separability parameter, formally, 
\begin{align*} 
    \delta:=\min_{i \neq j} \{ \min\{ \| \delta_{i,j}^+ \|_2, \| \delta_{i,j}^- \|_2 \} \}.
\end{align*}
\end{definition}

\cite{syz21} has given a property of the minimal eigenvalue of the $\NTK$ of a shifted $\ReLU$ activated neural network.
\begin{lemma}[Lemma C.1 in \cite{syz21}]\label{lem:lemma_C.1_in_syz21}
Let $m$ be number of samples of $H^{\dis}$. As long as 
\begin{align*}
    m = \Omega(\lambda^{-1}n \log (n/\rho)),
\end{align*}
then
\begin{align*}
    \Pr[ \lambda_{\min} ( H^{\dis} ) \geq \frac{3}{4} \lambda ] \geq 1- \rho.
\end{align*}
\end{lemma}

Prior work (\cite{os20}) has shown the relation between the data separability of the input of a neural network and the eigenvalue of the Kernel. But their work focuses on unshifted $\ReLU$ activated neural network. For shifted $\ReLU$ activated neural network, \cite{syz21} provided a further generalization to the shifted Kernel matrix.
\begin{theorem}[Theorem F.1 in \cite{syz21}]\label{thm:sep}
Consider $n$ points $x_1,\dots,x_n \in \R^d$ with $\ell_2$-norm all equal to $1$, and consider a random variable $w\sim{\cal N}(0,I_d)$. Define matrix 
\begin{align*}
    X \in \R^{n\times d}=[x_1~\dots~x_n]^\top. 
\end{align*}
Suppose the data separability of the $n$ points is $\delta$ where $\delta < \sqrt{2}$. Let shift parameter $b\geq 0$. Recall the continuous Hessian matrix $H^{\cts}$ is defined by
\begin{align*} 
        &~H^{\cts}_{i,j} \\
    :=  &~\E_{w \sim \N(0,I)} [ x_i^\top x_j {\bf 1}_{ w^\top x_i \geq b, w^\top x_j \geq b } ], \forall (i,j)\in [n]\times [n].
\end{align*} 
Let $\lambda:=\lambda_{\min}(H^{\cts})$. Then $\lambda$ has the follow sandwich bound,
\begin{align*}
      \lambda \in [   \exp(-b^2/2) \cdot \frac{\delta}{100n^2} , \exp(-b^2/2) ].
\end{align*}

\end{theorem}

\section{Technical Overview} \label{sec:tech_ow}
Here, we describe the outline of the main ideas required to prove Theorem \ref{thm:main}.

\paragraph{Key Ideas} Our algorithm relies on two simple but powerful observations about training 2-layer neural networks: The first observation is that the Jacobian matrix of the loss function is \emph{sparse} -- When weights are initialized randomly (with appropriately chosen bias parameter $b$), the fraction of nonzero entries in the Jacobi matrix is small.
Let $c$ be some fixed constant in $[0.1,1]$. We show that there is a choice of the parameter $b$ ensuring simultaneously that\footnote{We refer the readers to Section~\ref{sec:app_time} for more details.} 
\begin{itemize}
    \item For every input $x_i$, there are only $O(m^{1-c})$ activated  neurons;  
    \item The loss of each iteration is still at most a half of the loss of the last iteration. 
\end{itemize}
Our second observation is that the \emph{positions} of the nonzero entries in Jacobian matrix do not change much. This can be seen using the ``gradient flow" equation (via Gauss-Newton method) 
$
W_{t+1} = W_t - J_t^\top g_t,
$
where 
$
g_t := \arg\min_g \| J_tJ_t^\top g_t - (f_t-y) \|_2
$. 
Since the Jacobian matrix is sparse, it is not hard to see that only a little fraction of the weights need to be modified, i.e., the change from $W_t$ to $W_{t+1}$ involves updating only a small number of entries.

These two observations suggests a natural ``binary-search" type algorithm for updating the weight matrix in \emph{sublinear time} $o(nmd)$ per iteration. 

\paragraph{Threshold search data structure} 
We design a dynamic data structure for detecting and maintaining the non-zero entries of the Jacobian matrix $J$ of the network loss, as it evolves over iterations. Notice that whether an entry of $J$ is nonzero is equivalent to whether the inner product of an input $x_i$ and a weight $w_j$ is larger than $b$ (hence $\phi(w_j^\top x_i) > 0$). 

Accordingly, for every input $x_i$ in a batch, our algorithm maintains a binary search tree $\mathcal{T}_i$ where each leaf stores the inner product of $x_i$ and a weight $w_j$, and every non-leaf node stores the the \emph{maximum} of the values of its two children. In this way, non-zero entries can be found by searching, in all the trees $\{\mathcal{T}_i\}_{i\in[n]}$, from root to leaf and ignoring the unnecessary branches.

To implement this process efficiently, our  data structure 
 needs to support the following three operations (See Section \ref{sec:ds} for the formal details): 
 (1) \bf Initialization. \rm Given input vectors $x_1,\cdots,x_n$ and weight vectors $w_1,w_2,\cdots,w_m$ as input, it constructs $n$ binary trees $\mathcal{T}_1,\ldots, \mathcal{T}_n$ as described above, in $O(mnd)$ time. (2) \bf Updating of weights. \rm Taken an index $j \in [m]$ and a target value $z$, it replaces $w_j$ by $z$ in $O(nd + n\log m)$ time, as if initializing it with $w_1,w_2,\cdots,w_{j-1},z,w_{j+1},\cdots,w_m$ from scratch. 
(3) \bf Threshold Search Query. \rm Given an index $i$ and a threshold $\tau$ as input, our data structure rapidly finds all the weights $w_j$ which satisfies $\langle x_i, w_j \rangle \ge \tau$ in $O(K_q \log m)$ time, where $K_q$ is the number of satisfied weights. They can be used to find the nonzero entries of the Jacobian matrix $J$.

\paragraph{A Fast DNN Training Algorithm} 
Using the above dynamic data structure, we design a fast neural network training algorithm (see Algorithm \ref{alg:ours_informal}) composed of initialization and the (dynamic) training process. At initialization, it initializes the weight vector $W_0$ randomly.

The training process consists of maintaining \emph{sparse-recovery} sketches \cite{ac06,ldfu13,ns19}, online regression, and implicit weight maintenance. The goal of the first two techniques is to efficiently solve the $t$-th iteration regression problem (cf. \cite{bpsw21})
$
g_t := \arg\min_g\|J_tJ_t^\top g - (f_t-y)\|, \label{eq:regression}
$. 
The idea of implicit-weight-maintenance (via our data structure) is to update weights using the information propagated by the loss function.

The details of these three tools can be summarized as follows:
\begin{itemize}
    \item {\bf Sketch maintenance} The goal of sketch computing is to eliminate the disastrous influence of the high dimension of $J_t^\top$ (it has $md$ rows) when solving regression problem in Eq. \eqref{eq:regression}. Roughly speaking, in sketch computing, we find a sketch matrix $S$ with far smaller rows than $J_t^\top$ such that for any $d$-dimensional vector $x$, $\|SJ_t^\top x\|_2$ is very close to $\|J_t^\top x\|_2$. We show that sketch computing runs in $o(mnd)$ time.
    \item {\bf Iterative regression solver} To speed-up the solution of the online regression problem \eqref{eq:regression}, we show how to implement the iterative Conjugate-Gradient solver (a-la \cite{bpsw21}) \emph{in sub-linear time}
    to find an approximate solution $g_t$ in time $o(mnd) + \wt{O}(n^3)$. We then prove that the (accumulated) approximation errors do not harm the convergence rate and precision in our analysis.
    \item {\bf Implicit weight maintenance} The goal of implicit weight maintenance is to update weights according to the outcome of the iterative regression solver. Updating a single weight can be done by calling \textsc{Update} once. With the result of iterative regression and the fact that only $m^{-c}$ (where $c$ is some fixed constant $c \in [0.1,1]$)
    fraction of entries of $J_t$ are nonzero, we show that our algorithm finishes the update of weights in $o(mnd)$ time.
\end{itemize}

The details can be found in the pseudocode of Algorithm \ref{alg:ours_formal}.

\section{Convergence Analysis} \label{sec:correctness}

We focus on the convergence of our training algorithm in this section and leave the proof of running time in Section \ref{sec:running_time}. Specifically, the goal of this section is to prove the following result, which implies that for the neural network randomly initialized at the beginning of our algorithm, the loss function converges linearly with high probability. This section only contains a proof sketch. For more detailed correctness analysis, we refer the readers to section~\ref{sec:app_analysis}.
Our main convergence result is the following:
\begin{theorem}[Formal version of Theorem \ref{thm:main}, the convergence part]\label{thm:correctness}
Let $m$ be the width of the $\mathsf{NN}$. 
If
\begin{align*}
    m = \Omega (\max \{ \lambda^{-4} n^4, \lambda^{-2} n^2d\log(16n/\rho) \} ),
\end{align*}
then there is a constant $c'>0$ so that our algorithm obtains
\begin{align*} 
    \| f_{t+1} - y \|_2 \leq 0.5 \cdot \| f_t - y \|_2.
\end{align*}
It holds with probability
\begin{align*} 
    1-\frac{5}{2}\rho-n^2 \cdot \exp( -m\cdot \min\{c'e^{-b^2/2},\frac{R}{10\sqrt{m}}\} ) 
\end{align*}
The randomness comes from two parts: the initialization of neural network and iterative algorithm itself.
\end{theorem}

\paragraph{Bounding the Function Value and Jacobian at the Initialization}

We provide a lemma which shows that, with random initialization, as long as the 2-layer $\mathsf{NN}$ is wide enough, the norm of weight matrix, the initial predicted value and the Frobenius norm of the initial Jacobi matrix are all not large with high probability. We defer its proof into Section \ref{sec:app_analysis}.

\begin{lemma}[Informal version of Lemma \ref{lem:shifted_initialization_formal}] 
\label{lem:shifted_initialization_informal}
Consider shifted $\ReLU$. Suppose $m$ is the width of neural network. If $m = \Omega( d \log (16n/ \rho) )$, then we have
\begin{itemize}
	\item $\|W_0\|_2 = O(\sqrt{m})$.
	\item $\max_{i \in [n]} | f ( W , x_i ) | = O(1)$.
	\item $\max_{i \in [n]} \|J_{W_0, x_i}\|_{F} = O(1)$.
\end{itemize}
holds with probability $1 - \rho/2$.
\end{lemma}

\paragraph{$G$ does not move much when $W$ does not move much}

We provide a lemma which proves that, as long as the 2-layer $\mathsf{NN}$ is wide enough, then with high probability that, for randomly initialized weights $W_0$, if $W_0$ changes to $W$ after a small change, then the Gram matrix $G_W$ will not move much and the minimal eigenvalue of $G_W$ will also not move much. And We leave its proof in Section \ref{sec:app_analysis}.

\begin{lemma}[Shifted Perturbation Lemma, informal version of Lemma \ref{lem:shifted_small_move_eigenvalue_formal}] \label{lem:shifted_small_move_eigenvalue_informal}
Consider shifted ReLU with $b$. Let $b \geq 0$. Let $R_0>0$.  
Suppose 
\begin{align*}
m \ge \Omega(1) \cdot \max\{ b^2R_0^2, n^2 R_0^2 \lambda^{-2}, n\lambda^{-1} \log(n/\rho) \},
\end{align*}
then with prob.  
\begin{align*} 
\ge 1-\rho - n^2 \cdot \exp \big( -m\cdot \min\{c'e^{-b^2/2},\frac{R_0}{10\sqrt{m}}\} \big),
\end{align*}
for {\em any} weight $W \in \R^{d \times m}$ satisfying $\max_{r \in [m] }\|w_r - w_r(0)\|_2 \leq R_0/\sqrt{m}$, the following holds:
\begin{align*}
\|G_W - G_{W_0}\|_{F} \le \lambda / 2, ~~\text{and}~~
\lambda_{\min}(G_W) \ge \lambda / 2.
\end{align*}
Note that $w_r$ is representing the $r$-th column of $W$.
\end{lemma}

\paragraph{Perturbed weights difference under shifted NTK}

We give a lemma which proves that, as long as the 2-layer $\mathsf{NN}$ is wide enough, then with high probability that, for randomly initialized weights $W_0$, if $W_0$ changes to $W$ after a small change, the each row $J_{W,x_i}$ of Jacobi matrix $J_W$ will not change much, and the Frobenius norm of $J_W$ will also not change much. We leave its proof in Section \ref{sec:app_analysis}.  

\begin{lemma}[Informal version of Lemma  \ref{lem:small-move_formal}]
	\label{lem:small-move_informal}
	Suppose $R_0 \geq 1$ and $m = \tilde{\Omega}(n^2 R_0^2)$. With probability at least $1 - \rho$ over the random initialization of $W_0$, the following holds for {\em any} set of weights $w_1, \ldots w_m \in \R^{d}$ satisfying $\max_{r \in [m]} \|w_r - w_r(0)\|_2 \leq R_0/\sqrt{m}$,
	\begin{itemize}
		\item $\|W - W_0\| = O(R_0)$,
		\item $\|J_{W, x_i} - J_{W_0, x_i}\|_{2} = \tilde{O}  ( {R_0^{1/2}} / {m^{1/4}} )$ and $\|J_{W} - J_{W_0}\|_{F} = \tilde{O} ( {n^{1/2} R_0^{1/2}} / {m^{1/4}} )$,
		\item $\|J_{W}\|_{F} = O(\sqrt{n})$.
	\end{itemize}
\end{lemma}

\paragraph{Induction Hypothesis}

Finally, we're ready to prove our major theorem, Theorem \ref{thm:correctness}. Note that we only need to prove the induction hypothesis described in definition \ref{def:correctness_hypothesis}, then Theorem \ref{thm:correctness} holds by mathematical induction. We divide the proof of this hypothesis into 2 parts and prove them in section \ref{sec:part_I} and section \ref{sec:part_II} respectively.

\begin{definition}[Induction hypothesis]\label{def:correctness_hypothesis}
Define $R_0\approx n/\lambda$. For any fixed $t$, if 
\begin{align*}
    &~\|f_{t} - y\|_2 \leq \frac{1}{2} \|f_{t-1} - y\|_2 \\\text{~~and~~} &~\max_{ r \in [m]} \| w_r(t) - w_r(0) \|_2 \leq R_0 / \sqrt{m}.
\end{align*}
Then we have
\begin{align*}
    &~\|f_{t+1} - y\|_2 \leq \frac{1}{2} \|f_{t} - y\|_2 \\\text{~~and~~} &~\max_{r \in [m]} \| w_r(t+1) - w_r(0) \|_2 \leq R_0 / \sqrt{m}.
\end{align*}
\end{definition}
Formally, we describe the process of proving this hypothesis by the following Lemma \ref{lem:proof_hypothesis}, and specific proof can be seen in Section \ref{sec:induction}.

\begin{lemma} \label{lem:proof_hypothesis}
    Suppose initial weights $W_0$ satisfies the restriction of Lemma \ref{lem:shifted_initialization_informal}, \ref{lem:shifted_small_move_eigenvalue_informal} and \ref{lem:small-move_informal}, then the induction hypothesis described in Definition \ref{def:correctness_hypothesis} holds. 
\end{lemma}

\begin{algorithm*}[!th]\caption{Our training algorithm, informal version of Algorithm~\ref{alg:ours_formal}}\label{alg:ours_informal}
\begin{algorithmic}[1]
\Procedure{OurAlgorithm}{$X$, $\epsilon$} 
    \State {Initialization Step:}  randomly pick $W(0)$,  $T \gets \log(1/\epsilon)$, create a data structure
    \State Iterative Step: start with $t = 1$

        \State \hspace{4mm} Step 1: Do the sketch computing, it forms matrix $S\in \R^{N \times n}$
        \State \hspace{4mm} Implicitly write down the Jacobian matrix $J_t \in \R^{n \times md}$
        \State \hspace{4mm} Choose sketch related parameters as Definition~\ref{def:sketch_parameters}
        \State \hspace{4mm} Find sketching matrix $S\in \R^{ s_{\sketch} \times md}$ of $J_t^\top$
        
        \State \hspace{4mm} Step 2 Run an iterative regression algorithm with small size problem (size reduced by sketch)
        \State \hspace{4mm} Find approximated solution $g_t$ of regression problem $\arg\min_g\| (J_tS^\top) (SJ_t^\top) g - (f_t-y) \|$
        \State \hspace{4mm} Step 3: Maintain the weight implicitly
        \State \hspace{4mm} Update the weights $W_t$ to $W_{t+1}$  
        \State \hspace{4mm} Update the TS data structure using $W_{t+1}$
        \State \hspace{4mm} Increment $t$ by $1$
\EndProcedure
\end{algorithmic}
\end{algorithm*}

\section{Running Time Analysis} \label{sec:running_time}
This section focuses on analyzing the running time of our algorithm. It will show that when $m$ is large enough, the $\mathsf{CPI}$ is $o(nmd)$. 
We first present Theorem \ref{thm:running_time_informal}, our main running time result of the paper. We then provide three lemmas (Lemma~\ref{lem:sketch_time}, Lemma~\ref{lem:regression_time} and Lemma~\ref{lem:backward_time}) to prove our main theorem.
For more proof details of the running time, we refer the readers to Section~\ref{sec:app_time}. For simplicity of presentation, we use $o(m)$ and $o(mnd)$ in this section. In Section~\ref{sec:app_time}, we explicitly compute time by $m^{1-\alpha}$ and $m^{1-\alpha} nd$ where $\alpha \in [0.01,1)$ is some fixed constant. Our main running time result is the following:
\begin{theorem}[The running time part of Theorem \ref{thm:main}]\label{thm:running_time_informal}
The cost per iteration ($\mathsf{CPI}$) of our algorithm is 
\begin{align*}
    o(mnd) + \wt{O}(n^3)
\end{align*} without using $\mathsf{FMM}$. The $\mathsf{CPI}$ of our algorithm is 
\begin{align*}
    o(mnd) + \wt{O}(n^\omega)
\end{align*}
with using $\mathsf{FMM}$. 
\end{theorem}
\begin{proof}
Combining Lemma \ref{lem:sketch_time_formal}, Lemma \ref{lem:regression_time_formal} and Lemma \ref{lem:backward_time_formal}, the computation time of each iteration is 
{\small
\begin{align*}
      & ~\tilde{O}( n^2m^{0.76}d ) + \tilde{O}( nm^{0.76}d + n^3 ) \\
      &~~~~~~~~+ O( n^2m^{0.76}(d+\log m) ) \\
    = & ~\tilde{O}( n^2m^{0.76}d + n^3 ). 
\end{align*}
}
And if using $\mathsf{FMM}$, similarly the running time is 
\begin{align*}
    \tilde{O}( n^2m^{0.76}d + n^\omega ).
\end{align*}
By Theorem \ref{thm:correctness}, we have:
The time to reduce the training loss to $\epsilon$ is 
\begin{align*}
    \tilde{O}( (n^2m^{0.76}d + n^3 ) \log(1/\epsilon)).
\end{align*}
Taking advantage of $\mathsf{FMM}$, the time is 
\begin{align*}
    \tilde{O}( (n^2m^{0.76}d + n^\omega ) \log(1/\epsilon)). 
\end{align*}
Further, for example, if $m = n^{c}$ where $c$ is some large constant, then 
\begin{align*}
    n^2 m^{0.76} d \leq nm^{1-\alpha}d
\end{align*}
where $\alpha \in [0.1,0.24)$. 
Hence the time of each iteration is 
\begin{align*}
    \tilde{O}( m^{1-\alpha}nd + n^3 ),
\end{align*}
and the time to reduce the training loss to $\epsilon$ is 
\begin{align*}
    \tilde{O}( (m^{1-\alpha}nd + n^3 ) \log(1/\epsilon)). 
\end{align*}
Taking advantage of $\mathsf{FMM}$, the time is 
\begin{align*}
    \tilde{O}( (m^{1-\alpha}nd + n^\omega )\log(1/\epsilon)). 
\end{align*}
Thus we complete the proof.
\end{proof}
\vspace{-3mm}
\paragraph{Sketch Computing.} 
We provide the choice of sketching parameters in the following definition and give a lemma that analyzes the running time of the sketch computing process in Algorithm \ref{alg:ours_formal} under these parameters. It will imply that sketch computing is sublinear in $m$. 

\begin{definition}[sketch parameters]\label{def:sketch_parameters}
We choose sketch parameters in the following ways:
{\small
\begin{align*}
    \epsilon_{\sketch} =&~ 0.1, \\
    \delta_{\sketch} =&~ 1/\poly(n), \\
    s_{\sketch} =&~ n \poly( \epsilon_{\sketch}^{-1} , \log(n/\delta_{\sketch}) )
\end{align*}
}
\end{definition}

\begin{lemma}[Step 1, sketch computing. Informal version of Lemma~\ref{lem:sketch_time_formal}]\label{lem:sketch_time}
The sketch computing process of Algorithm \ref{alg:ours_informal} (its formal version is Algorithm~\ref{alg:ours_formal}) runs in time
$
o(mnd).
$
\end{lemma}
\paragraph{Iterative regression.} We present a lemma that analyzes the running time of the iterative regression process in Algorithm \ref{alg:ours_formal}. It implies that the running time of the iterative regression is sublinear in $m$.
\begin{lemma}[Step 2, running time of iterative regression. Informal version of Lemma~\ref{lem:regression_time_formal}]\label{lem:regression_time}
The iterative regression of Algorithm \ref{alg:ours_informal} (its formal version is Algorithm~\ref{alg:ours_formal}) runs in  time
\begin{align*}
    O( o(mnd) \log(n/\delta) +  n^3).
\end{align*}
Taking advantage of $\mathsf{FMM}$, it takes time
\begin{align*}
    O( o(mnd) \log(n/\delta) + n^\omega),
\end{align*}
where $\omega$ is the exponent of matrix multiplication. Currently $\omega \approx 2.373$ \cite{w12}.
\end{lemma}
\vspace{-3mm}
\paragraph{Implicit weight maintenance.} We give a lemma that analyzes the running time of the implicit weight maintenance process in Algorithm \ref{alg:ours_formal}. It implies that implicit weight maintenance process is sublinear in $m$.

\begin{lemma}[Step 3, implicit weight maintenance. Informal version of Lemma~\ref{lem:backward_time_formal}]\label{lem:backward_time}

The implicit weight maintenance of Algorithm \ref{alg:ours_informal}(its formal version is Algorithm~\ref{alg:ours_formal}) runs in time 
\begin{align*}
o(nm) \cdot (d+\log m).
\end{align*}
\end{lemma}

\section{Conclusion} \label{sec:conclusion}
The computational cost of training massively overparametrized DNNs is posing a major scalability barrier to the progress of AI, and motivates rethinking the traditional SGD-based training algorithms.  
For a neural network with $m$ parameters and an input batch of  $n$ datapoints in $\R^d$, previous state-of-art \cite{bpsw21, zmg19} show that dramatically fewer iterations (epochs) $T_\epsilon$ can be achieved via second-order methods, albeit with $O(mnd + n^3)$ cost per iteration, i.e., 
\begin{align*}
    O(T_\epsilon \cdot mnd)
\end{align*}
overal time to reduce training error below $\epsilon$.
Our work proposes a simple yet powerful view of the gradient flow process on wide DNNs ($m = \poly(n)$), as a collection of \emph{slowly-changing binary search trees}, enabling the design of a training algorithm for 2-layer overparametrized DNNs in \emph{sublinear} cost-per-iteration, while enjoying the ultra-fast convergence rate of second-order (Gauss-Newton) methods, i.e., in total time 
\begin{align*}
    \tilde{O}(T_\epsilon + mnd)
\end{align*}
instead of the aforementioned 
\begin{align*}
    \tilde{O}(T_\epsilon \cdot mnd).
\end{align*}

While this paper introduces novel contributions to the field of deep neural networks (DNNs) by proposing a sublinear cost-per-iteration (CPI) training method, there are several limitations and potential directions for future research. For example, the core insight of this paper is the representation of DNNs as binary search trees. This alternative view of DNNs is a promising direction, but it may have limitations when dealing with other types of networks (for example, recurrent neural networks). 

Future research could explore adapting this approach to other network architectures. Whatsmore, the research currently focuses on two-layer ReLU neural networks. Future research could explore the applicability of the proposed training method to more complex, deeper architectures or different activation functions, thus broadening the scope of this work. 

We believe our new representation of DNNs will have further algorithmic implications to scaling training and inference in DNNs. 


\ifdefined\isarxiv
\else

\input{impact_statement}
\bibliography{ref}

\bibliographystyle{icml2024}
\fi


\onecolumn
\appendix


\section*{Appendix}

\paragraph{Roadmap.} The appendix of this paper is organized as follows. Section \ref{sec:app_preli} presents the preliminary tools which are used in the other parts of appendix. Section \ref{sec:ds} presents the complete description and implementation of the threshold search data structure. Section \ref{sec:app_alg} presents the formal algorithm representation of our fast neural network training algorithm. Section \ref{sec:app_analysis} shows the omitted proofs of some lemmas in the convergence analysis. Section \ref{sec:induction} presents the complete   the induction hypothesis proof to show the convergence of our training algorithm. Section \ref{sec:app_time} presents the formal proof of the running time of our training algorithm, especially shows a more specific conclusion compared with the main body. Section \ref{sec:combination} presents a formal analysis of the applying conditions of our training algorithm.

\section{Preliminary} \label{sec:app_preli}

This section shows some preliminary tools to be used later. 
In Section~\ref{sec:app_prel:fmm} we introduce the result for fast matrix multiplication. In Section~\ref{sec:app_prob} we state the probability tools to be used. In Section~\ref{sec:prel_perturb_w} we presented a previous result on the relationship of changes of weights and the change of the shifted NTK matrix. In Section~\ref{sec:fast-regression} we provide some useful results about fast regression. In Section~\ref{sec:quantum_benefit_assumptions} we provide results about sparsity-based preserving. 


\subsection{Fast matrix multiplication}
\label{sec:app_prel:fmm}
We state a standard fact for fast matrix multiplication ($\mathsf{FMM}$).
\begin{fact}[$\mathsf{FMM}$]
Given an $n \times n$ matrix $A$ and another $n \times n$ matrix $B$, the time of multiplying $A$ and $B$ is $n^{\omega}$, where $\omega \approx 2.373$ is the exponent of matrix multiplication. Currently, $\omega \approx 2.373$ \cite{w12}.
\end{fact}

\subsection{Probability Tools}
\label{sec:app_prob}

We list some probability tools which are useful in our analysis.

\begin{lemma}[Chernoff bound \cite{c52}]\label{lem:chernoff}
Let $Z = \sum_{i=1}^n Z_i$, where $Z_i=1$ with probability $p_i$ and $Z_i = 0$ with probability $1-p_i$, and all $Z_i$ are independent. We define $\mu = \E[Z] = \sum_{i=1}^n p_i$. Then \\
1. $ \Pr[ Z \geq (1+\delta) \mu ] \leq \exp ( - \delta^2 \mu / 3 ) $, $\forall \delta > 0$ ; \\
2. $ \Pr[ Z \leq (1-\delta) \mu ] \leq \exp ( - \delta^2 \mu / 2 ) $, $\forall  \delta \in (0,1)$. 
\end{lemma}

\begin{lemma}[Hoeffding bound \cite{h63}]\label{lem:hoeffding}
Let $Z_1, \cdots, Z_n$ denote $n$ independent bounded variables in $[a_i,b_i]$. Let $c_i = (b_i-a_i)$ Let $Z = \sum_{i=1}^n Z_i$, then we have
\begin{align*}
\Pr[ | Z - \E[Z] | \geq t ] \leq 2\exp \left( - \frac{2t^2}{ \sum_{i=1}^n c_i^2 } \right).
\end{align*}
\end{lemma}

\begin{lemma}[Anti-concentration inequality]
\label{lem:anti_gaussian}
Let $Z \sim {\cal N}(0,\sigma^2)$,
that is,
the probability density function of $Z$ is given by $\phi(x)=\frac 1 {\sqrt{2\pi\sigma^2}}e^{-\frac {x^2} {2\sigma^2} }$.
Then
\begin{align*}
    \Pr[|Z|\leq t] \leq \frac{4}{5} \frac{t}{\sigma}. 
\end{align*}
\end{lemma}

\subsection{Perturbed $w$ for Shifted NTK}
\label{sec:prel_perturb_w}
We present a lemma from previous work in \cite{syz21}. They show that in general, small changes of weights only lead to small change of the Shifted NTK matrix.

\begin{lemma}[Lemma C.2 in \cite{syz21}, perturbed $w$ for shifted NTK]\label{lem:boundary_of_G}
Suppose $b>0$. Assume $R\leq 1/b$. Suppose $m = \Omega( \lambda^{-1} n \log(n/\rho) )$. Define function $H$ which maps $\R^{m \times d}$ to $\R^{n \times n}$ as follows:
\begin{align*}
   \text{~the~$(i,j)$-th~entry~of~} H(W) \text{~is~} \frac{1}{m} x_i^\top x_j \sum_{r=1}^m {\bf 1}_{ w_r^\top x_i \geq b, w_r^\top x_j \geq b } .
\end{align*} 
Let $m$ vectors $w_1, w_2, \cdots, w_m$ sampled from ${\N}(0,I_d)$ and let $\wt{W} = [w_1 ~ w_2 ~ \cdots ~ w_m]$. Then there exist constants $c>0$ and $c'>0$ such that, for all $W \in \R^{d \times m}$ with $\| \wt{W} - W \|_{\infty,2} \leq R$, the following holds:
\begin{itemize}
    \item Part 1, $\|  H(\wt{W}) - H (W) \|_F \leq n \cdot \min\{ c\cdot \exp(-b^2/2), 3R\}$ holds with prob.  $\ge 1-n^2\cdot \exp(-m\cdot \min\{c'\cdot  \exp(-b^2/2), R/10\})$.
    \item Part 2, $\lambda_{\min}(H(W))\geq \frac{3}{4}\lambda - n \cdot \min\{ c\cdot \exp(-b^2/2), 3R\}$ holds with prob. $\ge 1-n^2\cdot \exp(-m\cdot \min\{c'\cdot  \exp(-b^2/2), R/10\})-\rho$.
\end{itemize}
\end{lemma}

\subsection{Fast regression solver}
\label{sec:fast-regression}
We list some useful conclusions about fast regression from \cite{bpsw21}.

\begin{lemma}[Lemma B.2 in \cite{bpsw21}]
	\label{lem:regular-regression}
	Consider the the regression problem 
	\begin{align*}
	\min_{x}\|Bx - y\|_2^2.
	\end{align*}
	Suppose $B$ is a PSD matrix with $\frac{3}{4} \leq\|Bx\|_2 \leq \frac{5}{4}$ holds for all $\|x\|_2 = 1$. Using gradient descent, after $t$ iterations, we obtain
	\begin{align*}
	\|B(x_t - x^{\star})\|_2 \leq c^{t} \cdot \|B(x_0 - x^{\star})\|_2
	\end{align*}
	for some constant $c \in (0,0.9]$.
\end{lemma}

\begin{lemma}[Lemma B.1 in \cite{bpsw21}]\label{lem:fast-regression}
	Suppose there is a matrix $Q \in \R^{N \times k}$ ($N \geq k\poly(\log k)$), with condition 
	number $\kappa$ (i.e., $\kappa= \sigma_{\max}(Q) / \sigma_{\min}(Q)$), consider this minimization problem
	\begin{align}
	\label{eq:reg1}
	\min_{x \in \R^k} \| Q^{\top} Q x - y \|_{2}.
	\end{align}
	It is able to find a vector $x'$ 
	\begin{align*}
	\| Q^{\top} Q x' - y\|_{2} \leq \|y\|_2 \cdot \epsilon
	\end{align*}
	in $
	    {\cal T}_{\mathrm{precond}} + {\cal T}_{\mathrm{iters}} \cdot {\cal T}_{\mathrm{cost}}
	$ time
	where
	\begin{itemize}
	    \item ${\cal T}_{\mathrm{precond}} = \wt{O}(Nk+k^3)$ without using $\mathsf{FMM}$, $\wt{O}(Nk+k^{\omega})$ using $\mathsf{FMM}$.
	    \item ${\cal T}_{\mathrm{iters}} = O(\log(\kappa/\epsilon))$, 
	    \item ${\cal T}_{\mathrm{cost}} = \wt{O}(Nk)$. 
	\end{itemize}
\end{lemma}

The above lemma and preconditioning property implies that the iterative regression will take $\log(\kappa/\epsilon)$ iterations.
\begin{corollary} \label{cor:iterative_regression}
    Solving regression problem \eqref{eq:reg1} needs $O(\log(\kappa/\epsilon))$ iterations using the above method.
\end{corollary}
The cost per iteration in the iterative regression is too slow for our application. In Section~\ref{sec:app_time}, we will show how to improve the cost per iteration while maintaining the same number of iterations.

\subsection{Sparsity-based Preserving} \label{sec:quantum_benefit_assumptions}
We present a tool from the paper \cite{syz21}. Firstly, we provide a definition.

\begin{definition}
For every $t \in \{0,1,\cdots,T\}$.
For every $i \in [n]$. We use ${\cal S}_{i,\mathrm{fire}}(t) \subset [m]$ to represent the set of neurons that are ``fire'' at time $t$, i.e.,
\begin{align*}
{\cal S}_{i,\mathrm{fire}}(t) := \{ r \in [m] : \langle w_r(t), x_i \rangle  > b \}.
\end{align*}
For all $t \in \{0,1,\cdots,T\}$, define $k_{i,t} := |{\cal S}_{i,\mathrm{fire}}(t) |$ to express the number of fire neurons for $x_i$.
\end{definition}

The following lemma (Lemma 3.8 in \cite{syz21}) show that with the increase of the shifted paramater, the initial neural network will become sparser.
\begin{lemma}[Sparsity preserving] \label{lem:sparse_initial}
Assume $m$ is number of neurons.
For shifted parameter $b > 0$, if we use $\phi_b$ as the activation function of a 2-layer neural network, then after initialization, with prob. $\ge 1-n\cdot \exp(-\Omega(m \cdot \exp(-b^2/2)))$, we have for every $i$, $k_{i,0}$ is not larger than $O(m\cdot \exp(-b^2/2))$.
\end{lemma}

Using the above lemma, we can obtain the following result,
\begin{corollary}[
] \label{cor:sparse}
If we set shifted parameter $b = \sqrt{0.48\log m}$ then $k_0 = m^{0.76}$. 
For $t= m^{0.76}$, 
\begin{align*}
    \Pr \left[ |{\cal S}_{i,\mathrm{fire}}(0)| > 2m^{0.76} \right] \leq \exp \big(- \min \{ m R, O(m^{0.76} ) \} \big) .
\end{align*}
\end{corollary}

\section{Threshold search data structure} \label{sec:ds}

This section gives a data structure which can efficiently find all the weights $w_j$ such that $\langle w_j, x_i \rangle \ge \tau$ for each given input $x_i$ and real number $\tau$. Specifically,  Section \ref{sec:main} formally proposes this data structure.   Section \ref{sec:init} proves the running time of \textsc{Init} satisfies the requirement of Theorem \ref{thm:tree_ds}.   Section \ref{sec:update} proves the running time of \textsc{Update} satisfies the requirement of Theorem \ref{thm:tree_ds}.  Section \ref{sec:query1} proves the running time of \textsc{Query} satisfies the requirement of Theorem \ref{thm:tree_ds}.  Section \ref{sec:query2} proves the correctness of \textsc{Query} in Theorem \ref{thm:tree_ds}.

\subsection{Main result} \label{sec:main}
In this section, we are going to present our key theorem (Theorem~\ref{thm:tree_ds}).
\begin{theorem}[Our tree data structure]\label{thm:tree_ds}
There exists a data structure which requires $O(mn + nd + md)$ spaces and supports the following procedures:
\begin{itemize}
    \item \textsc{Init}$(\{w_1,w_2, \cdots, w_m\} \subset \R^d, \{x_1, x_2, \cdots, x_n\} \subset \R^d$. Given a series of weights $w_1,w_2,\cdots,w_m$ and datas $x_1, x_2, \cdots, x_n$, it preprocesses in time $O(mnd)$.
    \item \textsc{Update}$(z \in \R^d,j\in [m])$. Given a new weight vector $z \in \R^d$ and index $j \in [m]$, it updates weight $w_j$ with $z$ in time $O(n (d+ \log m))$. 
    \item \textsc{Query}$(i \in [n], \tau \in \R)$. Given a query index $i \in [n]$ and a threshold $\tau \in \R$, it finds all index $j \in[m]$ such that $\langle w_j, x_i \rangle \geq \tau$ in time $O(K_q \cdot \log m)$, where $K_q:= | \{ j \in [m] ~|~ \langle w_j , x_i \rangle \geq \tau \} |$.
\end{itemize}
\end{theorem}

\begin{proof}
Since $W$ takes $O(md)$ space, $X$ takes $O(nd)$ space, each binary tree $T_i$ stores $O(m)$ data, the data structure uses $O(mn+nd+md)$. Then we use the following Lemma \ref{lem:init_time}, \ref{lem:update_time}, \ref{lem:query_time} and \ref{lem:query_correct} to prove the correctness and running time of this data structure.
\end{proof}

\begin{algorithm}[!ht]\caption{Our tree data structure: members, init}\label{alg:tree_ds_init}
\begin{algorithmic}[1]
\State {\bf data structure} \textsc{Tree} \Comment{Theorem~\ref{thm:tree_ds}}
\State {\bf members}
\State \hspace{4mm} $W \in \R^{m\times d}$ ($m$ weight vectors)
\State \hspace{4mm} $X \in \R^{n \times d}$ ($n$ data points)
\State \hspace{4mm} Binary tree $T_1, T_2, \cdots, T_n$ \Comment{We create $n$ binary search trees, each tree uses $O(mn)$ space}
\State {\bf end members}
\State
\State {\bf public:}
\Procedure{Init}{$w_1,w_2, \cdots, w_m \in \R^d, x_1, x_2, \cdots, x_n \in \R^d$} \Comment{Lemma~\ref{lem:init_time}}
    \For{$i=1 \to n$} \label{init:for1}
        \State $x_i \gets x_i$
    \EndFor
    \For{$j=1 \to m$} \label{init:for2}
        \State $w_j \gets w_j$
    \EndFor
    \For{$i=1 \to n$} \label{init:for3}\Comment{for data point, we create a tree}
        \For{$j=1 \to m$} 
            \State $u_j \gets \langle x_i, w_j \rangle$ \label{init:u_j}
        \EndFor
        \State $T_i \gets \textsc{MakeBinarySearch}(u_1, \cdots, u_m)$ \label{init:T_i}
        \State \Comment{Each node stores the maximum value for his two children}
    \EndFor
\EndProcedure
\State {\bf end data structure}
\end{algorithmic}
\end{algorithm}

\begin{algorithm}[!ht]\caption{Our dynamic data structure: update}\label{alg:tree_ds_update}
\begin{algorithmic}[1]
\State {\bf data structure} \textsc{Tree} \Comment{Theorem~\ref{thm:tree_ds}}
\State {\bf public:}
\Procedure{Update}{$z\in\R^d, j \in [m]$} \Comment{Lemma~\ref{lem:update_time}}  
\State $w_j \gets z$ 
\For{$i \in [n] $} \label{update:for}
    \State $l \gets$ the $j$-th leaf of tree $T_i$
    \State $l.\text{value} \gets \langle z, x_i \rangle$ \label{update:l.value}
    \While{$l$ is not root}
        \State $p$ $\gets$ parent of $l$
        \State $a$ $\gets$ left child of $p$
        \State $b$ $\gets$ right child of $p$
        \State $p.\text{value} \gets \max \{ a.\text{value}, b.\text{value} \}$
        \State $l \gets p$
    \EndWhile
\EndFor
\EndProcedure
\State {\bf end data structure}
\end{algorithmic}
\end{algorithm}

\begin{algorithm}[!ht]\caption{Our dynamic data structure: query}\label{alg:tree_ds_query}
\begin{algorithmic}[1]
\State {\bf data structure} \textsc{Tree} \Comment{Theorem~\ref{thm:tree_ds}}
\State {\bf public:}
\Procedure{Query}{$i \in [n], \tau \in \R_{\geq 0}$} \Comment{Lemma~\ref{lem:query_time}}  
\State \textsc{QRecursive}($\tau,\mathrm{root}(T_i)$)
\EndProcedure
\State  
\State {\bf private:}  
\Procedure{QRecursive}{$\tau \in \R_{\geq 0}, r\in T$}
\If{$r$ is leaf}
\If{$r$.value $>\tau$}
\State \Return $r$.index
\EndIf
\Else
\State $r_1\gets$ left child of $r$, $r_2\gets$ right child of $r$
\If{$r_1.\text{value} \geq \tau$}
    \State $S_1 \gets $\textsc{QRecursive}$(\tau,r_1)$
\EndIf
\If{$r_2.\text{value} \geq \tau$}
    \State $S_2 \gets $\textsc{QRecursive}$(\tau,r_2)$
\EndIf
\EndIf
\State \Return $S_1 \cup S_2$
\EndProcedure
\State {\bf end data structure}
\end{algorithmic}
\end{algorithm}

\subsection{Running Time of Init} \label{sec:init}
We prove Lemma~\ref{lem:init_time}, which presents the running time for the \textsc{Init} operation. The corresponding algorithm is shown in Algorithm~\ref{alg:tree_ds_init}.
\begin{lemma}[Running time of \textsc{Init}]\label{lem:init_time}
Given a series of weights $\{w_1,w_2,\cdots,w_m\}\subset\R^d$ and datas $\{x_1, x_2, \cdots, x_n\}\subset\R^d$, the procedure \textsc{Init} (Algorithm~\ref{alg:tree_ds_init}) preprocesses in time $O(nmd)$.
\end{lemma}
\begin{proof}
The \textsc{Init} consists of two independent for loops and two recursive for loopss. The first for loop (start from line~\ref{init:for1}) has $n$ iterations, which takes $O(nd)$ time. The second for loop (start from line~\ref{init:for2}) has $m$ iterations, which takes $O(md)$ time.
Now we consider the recursive for loop. The outer loop (line~\ref{init:for3}) has $n$ iterations. In inner loop has $m$ iterations. In every iteration of the inner loop, line~\ref{init:u_j} runs in $O(d)$ time. Line~\ref{init:T_i} takes $O(m)$ time.
Putting it all together, the  \textsc{Init} runs in time
\begin{align*}
     & ~ O(nd+md+n(md+m))\\
    =& ~ O(nmd)
\end{align*}
So far, the proof is finished.
\end{proof}

\subsection{Running Time of \textsc{Update}} \label{sec:update}
We prove Lemma~\ref{lem:update_time}. The corresponding algorithm is shown in Algorithm~\ref{alg:tree_ds_update}.
\begin{lemma}[Running time of \textsc{Update}]\label{lem:update_time}
Given a weight $z\in\R^d$ and index $j\in [m]$, the procedure \textsc{Update} (Algorithm~\ref{alg:tree_ds_update}) updates weight $w_j$ with $z$ in  $O(n\cdot(d+\log m))$ time.
\end{lemma}
\begin{proof}
The time of \textsc{Update} mainly comes from the forloop (line~\ref{update:for}), which consists of $n$ iterations. In each iteration, line~\ref{update:l.value} takes $O(d)$ time, and the while loop takes $O(\log m)$ time since it go through a path bottom up.
Putting it together, the running time of \textsc{Update} is $O(n(d+\log m))$.
\end{proof}

\subsection{Running Time of Query} \label{sec:query1}

We prove Lemma~\ref{lem:query_time}, which is the running time for the \textsc{Query} operation. The corresponding algorithm is shown in Algorithm~\ref{alg:tree_ds_query}.
\begin{lemma}[Running time of \textsc{Query}]\label{lem:query_time}
Given a query index $i \in [n]$ and a threshold $\tau > 0$, the procedure \textsc{Query} (Algorithm~\ref{alg:tree_ds_query}) runs  in time $O(K_q \cdot \log m)$, where $K_q := | \{j \in [m] : \langle w_j ,  x_i \rangle >\tau\} |$.
\end{lemma}
\begin{proof}
The running time comes from \textsc{QRecursive} with input $\tau$ and $\mathrm{root}(T_i)$. In \textsc{QRecursive}, we start from the root node $r$ and find indices in a recursive way. The \textsc{Init} guarantees that for a node $r$ satisfying $r.\mathrm{value} > \tau$, the sub-tree with root $r$ must contains a leaf whose value is greater than $\tau$ If not satisfied, all the values of the nodes in the sub-tree with root $r$ is less than $\tau$. This guarantees that all the paths it searches do not have any branch that leads to unnecessary leaves.
Our data structure will report all the indices $i$ satisfying $\langle w_i,q\rangle>\tau$. Since the depth of $T$ is $O(\log m)$, the running time of \textsc{Query} is $O|K_q|\cdot\log m)$.
\end{proof}

\subsection{Correctness of Query} \label{sec:query2}
We prove Lemma~\ref{lem:query_correct}, which shows the correctness for the \textsc{Query} operation.
\begin{lemma}[Correctness of \textsc{Query}]\label{lem:query_correct}
Given a query index $i \in [n]$ and a threshold $\tau > 0$, the procedure \textsc{Query} (Algorithm~\ref{alg:tree_ds_query}) finds all index $j\in[m]$ such that $\langle x_i, w_j \rangle >\tau$.
\end{lemma}
\begin{proof}
Fix $i\in[n]$, for all $j\in[m]$, suppose the $j$-th leaf of $T_i$ is $l$, the root of $T_i$ is $r$, and the path from $r$ to $l$ is 
\begin{align*}
r=p_0\to p_1\to\cdots \to p_k=l.
\end{align*}
If $\langle x_i,w_j \rangle > \tau$, first $j\in$ \textsc{QRecursive}($p_k$), then, suppose $j\in$ \textsc{QRecursive}($p_{t+1}$), then $p_{t+1}$.value $\ge \langle w_j,x_i \rangle>\tau$, thus $j\in$ \textsc{QRecursive}($p_{t+1}$) $\subseteq$ \textsc{QRecursive}($p_t$). Hence by induction, $j\in$ \textsc{QRecursive}($p_0$)=\textsc{Query}($i,\tau$). 
If $\langle x_i,w_j \rangle \le \tau$, since $l$.value $\ge \tau$, $j$ will not be returned.
Thus \textsc{Query} finds exactly all the index $j\in[m]$ such that $\langle x_i,w_j \rangle > \tau$.

\end{proof}

\section{Formal Algorithm Representation} \label{sec:app_alg}
We have given a concise representation of our training algorithm (Algorithm \ref{alg:ours_informal}) in previous sections, for facilitating understanding. For the sake of completeness and convenient implementation, this section gives a formal algorithm representation of our fast neural network training algorithm. (See Algorithm \ref{alg:ours_formal}.) 

This algorithm starts with initializing weights $W_0$ and setting shifted parameter $b$. After that, it repeatedly executes sketch computing, iterative regression and implicit weight maintenance until enough times. Specifically, sketch computing computes a sketch matrix $S$ for $J_t^\top$ with property $\| SJ_t^\top x \|$ is closed to $\| J_t^\top x \|$ for every $x$ with large probability. Iterative regression makes use of a fast regression solver to find an approximate solution of 
\begin{align*}
g_t := \arg\min_g\|J_tJ_t^\top g - (f_t-y)\| 
\end{align*}
with the help of the sketch matrix $S$.

Implicit weight maintenance utilizes the threshold search data structure to update weights using the information propagated by the iterative regression.

\begin{algorithm}[!th]\caption{Our training algorithm, Formal version of Algorithm~\ref{alg:ours_informal}}\label{alg:ours_formal}
\begin{algorithmic}[1]
\Procedure{OurAlgorithm}{$\{x_i\}_{i \in [n]}$, $\epsilon$} 
    \State {\color{blue}/*Initialization*/}
    \State Randomly pick $W(0)$ \label{row:W_0}
    \State \textsc{Tree}.\textsc{Init}($\{ (W_0)_r \}_{r \in [m] }, m, \{ x_i \}_{i\in [n]}, n, d$) \label{row:init}
    \Comment{Alg.~\ref{alg:tree_ds_init}}
    \State $T \gets \log(1/\epsilon)$, $b \gets \sqrt{0.48\log m}$
    \State {\color{blue}/*Iterative Algorithm*/}
    \For{$t=1 \to T$} \label{row:mainloop}
        \State {\color{blue}/*Three computation tasks*/}
        \State {\color{blue}/*Step 1, Sketch computing*/}
        \State Implicitly write down the Jacobian matrix $J_t \in \R^{n \times md}$ \label{line:sketch_start}
        \State Let $A = J_t^\top$
        \State $\epsilon_{\sketch} \gets 0.1$
        \State $\delta_{\sketch} \gets 1/\poly(n)$
        \State $s_{\sketch} \gets n \poly( \epsilon_{\sketch}^{-1} , \log(n/\delta_{\sketch}) )$
        \State Find sketching matrix $S\in \R^{ s_{\sketch} \times md}$ of $A$
        \For{$i=1 \to n$}
            \State $Q_i \gets \textsc{Tree}.\textsc{Query}(i,b)$ \Comment{$Q_i \subset [m]$} 
            \State \Comment{ Theorem~\ref{thm:neural_network_sparisity} implies $|Q_i| = O(m^{0.76})$}
            \State Let $D_i \in \R^{m\times m}$ denote a matrix where $(D_i)_{j,j}=1$ if $j \in Q_i$
            \State Let $D_i \otimes I_d$ denote an $md \times md$ matrix
            \State $B_{*,i} \gets S \cdot (D_i \otimes I_d) \cdot A_{*,i} $ \Comment{$S$ is a sketching matrix}
        \EndFor
        \State Let $Q = \cup_i Q_i$ \label{line:sketch_end}
        \State Let $D$ denote the diagonal version of $Q$
        \State {{\color{blue}/*Step 2, Iterative regression*/}}
        \State Compute $R \in \R^{n \times n}$ such that $SAR$ has orthonormal columns via QR decomposition \label{row:QR_decomposition} \label{line:iter_start}
        \State $\tau \gets 1 $
        \State Compute $f_t$ based on $Q$
        \State Compute $y_{\mathrm{reg}} \gets f_t - y$
        \State $\epsilon_{\text{reg}} \gets \frac{1}{6}\sqrt{\frac{\lambda}{n}}$ \label{row:eps_reg}
        \While{$ \| A^\top (D \otimes I_d) A R z_t - y_{\mathrm{reg}} \|_2 \geq \epsilon_{\mathrm{reg}} $} \label{row:iterative_regression}
            \State $z_{t+1} \gets z_t - (R^\top A^\top (D \otimes I_d) A R)^\top (R^\top A^\top (D \otimes I_d) A R z_t - R^\top y_{\mathrm{reg}})$
            \State $\tau \gets \tau+1$
        \EndWhile
        \State Compute $g_t \gets z_t$ \label{line:iter_end}
        \State {\color{blue}/*Step 3, Implicit weight maintenance*/}
        \State {\color{blue} /* $W_{t+1} \gets W_t - J_t^\top g_t$ */}
        \State Let $K \subset[m]$ denote the set of coordinates, we need to change the weights \label{line:maintenance_start}
        \State \Comment{Theorem~\ref{thm:neural_network_sparisity} implies $|K| = O(m^{0.76}n)$}
        \For{$r \in K$} \label{row:backward_computation}
            \State Compute $(W_{t+1})_r$ \Comment{$(W_{t+1})_r \in \R^d$}
            \State \textsc{Tree}.\textsc{Update}$( ( W_{t+1} )_r , r )$ \Comment{Alg.~\ref{alg:tree_ds_update}}
        \EndFor \label{line:maintenance_end}
    \EndFor
\EndProcedure
\end{algorithmic}
\end{algorithm}

\section{More Details about Convergence Analysis} \label{sec:app_analysis}

The convergence analysis is shown in Section \ref{sec:correctness}. It uses Lemma \ref{lem:shifted_initialization_informal}, Lemma \ref{lem:shifted_small_move_eigenvalue_informal} and Lemma \ref{lem:small-move_informal} without proofs. In this section, we formally present the proofs of the three lemmas. In Section~\ref{sec:app_analysis:lem_shifted_initialization_formal}, we provide the proof of Lemma~\ref{lem:shifted_initialization_informal}. In Section~\ref{sec:app_analysis:lem_shifted_small_move_eigenvalue_informal}, we provide the proof of Lemma~\ref{lem:shifted_small_move_eigenvalue_informal}. In Section~\ref{sec:app_analysis:lem_small-move_informal}, we provide the proof of Lemma~\ref{lem:small-move_informal}. 

\subsection{Proof of Lemma~\ref{lem:shifted_initialization_informal}}\label{sec:app_analysis:lem_shifted_initialization_formal}

\begin{lemma}[Formal version of Lemma \ref{lem:shifted_initialization_informal}] 
\label{lem:shifted_initialization_formal}
For 2-layer ReLU activated neural network, suppose $m = \Omega( d \log (16n/ \rho) )$, then the following 
\begin{itemize}
	\item $\|W_0\|_2 = O(\sqrt{m})$.
	\item $| f ( W , x_i ) | = O(1)$, for $i \in [n]$. 
	\item $\|J_{W_0, x_i}\|_{F} = O(1)$, for $i \in [n]$.
\end{itemize}
holds with prob. $\ge 1 - \rho/2$.
\end{lemma}

\begin{proof}
(a) The first term can be seen in Corollary 5.35 of \cite{v10}. Notice that $W_0\in\R^{m\times d}$ is a
Gaussian random matrix, the Corollary gives 
\begin{align*}
   \Pr[ \|W_0\|_2 \le \sqrt{m} + \sqrt{d} + t ] \geq 1-2e^{-\frac{t^2}{2}}.
\end{align*}
Let us set $m=\max\{ d, \sqrt{2\log(8/\rho)} \}$, it gives $\|W_0\|_2 \le 3\sqrt{m}$ with probability $1-\rho/4$.

(b) For the second term, first, $a_r,r\in[m]$ are Rademacher variables, thereby 1-sub-Gaussian, so with probability $1-2e^{-mt^2/2}$ we have $\frac{1}{m}|\sum_{r=1}^m a_r|\le t$. This means if we take $m=\Omega( \log(16/\rho) )$, 
\begin{align}
    \Pr[\frac{1}{\sqrt{m}}\sum_{r=1}^m a_r = O(1)] \ge 1-\frac{\rho}{8}. \label{eq:initial_1}
\end{align}
Next, the vector $v_i = W_0^{\top}x_i \in \mathbb{R}^{m}$ is standard Gaussion vector. Write $a = \left[ \begin{array}{cccc}
    a_1 & a_2 & \cdots & a_m 
\end{array} \right]^{\top}$, since activation function $\phi_b$ is $1$-Lipschitz, with a vector $a$ fixed, the function 
\begin{align*}
    \Phi:\mathbb{R}^m\to \mathbb{R}, v_i \mapsto \frac{1}{\sqrt{m}}a^{\top} \phi_b(v_i) = f(W_0,x_i)
\end{align*}
has a Lipschitz parameter of $1/\sqrt{m}$. 

Due to the concentration of a Lipschitz function under Gaussian variables (Theorem 2.26 in \cite{w19}), 
\begin{align*}
    \Pr[|\Phi(v_i)-\mathbb{E}_{W_0}(\Phi(v_i))| \ge t] \le 2 e^{-\frac{mt^2}{2}},
\end{align*}
which means if $m=\Omega(\log(16n/\rho))$,
\begin{align}
    \left| \frac{1}{\sqrt{m}}a^{\top}\phi_b(W_0x_i) - \frac{1}{\sqrt{m}}\left(\sum_{r \in [m] }a_r\right) \E_{w\sim {\cal N}(0,I_d) }[\phi_b( \langle w,x_i \rangle )] \right| = O(1) \label{eq:initial_2}
\end{align}
holds at the same time for all $i\in[n]$ with probability $1-\frac{\rho}{8}$.

We know 
\begin{align}
    |\mathbb{E}_{w\sim {\cal N}(0,I_d) }[\phi_b(wx_i)]| \le |\phi_b(0)| + \mathbb{E}_{\xi\sim {\cal N}(0,1)}[|\xi|] = O(1). \label{eq:initial_3}
\end{align}
Plugging in Eq. \eqref{eq:initial_1}, \eqref{eq:initial_3} into Eq. \eqref{eq:initial_2}, we see that once $m=\Omega( \log(16n/\rho) )$, then with probability $1-\rho/4$, for all $i\in[n]$, 
\begin{align*}
    | f(W_0,x_i) | = | \frac{1}{\sqrt{m}}a^{\top} \phi_b(W_0x_i) | = O(1).
\end{align*}
(c) Let $d_{W,x}=\phi_b'(Wx)$ denote the element-wise derivative of the activation function, since $\phi_b$ is $1$-Lipschitz, we have $\|d_{W,x}\|_{\infty}=O(1)$. Note that $J_{W,x} = \frac{1}{\sqrt{m}}( (d_{W,x} \circ a) x^{\top} )$ where $\circ$ denotes the element-wise product, we can easily know \begin{align*}
\|J_{W,x_i}\|_F \le \frac{1}{\sqrt{m}} \cdot \|\text{Diag}(d)\|_2 \cdot  \|a\|_2 \cdot \|x\|_2 = O(1).
\end{align*}
\end{proof}

\subsection{Proof of Lemma~\ref{lem:shifted_small_move_eigenvalue_informal}}
\label{sec:app_analysis:lem_shifted_small_move_eigenvalue_informal}

\begin{lemma}[Shifted Perturbation Lemma, formal version of Lemma \ref{lem:shifted_small_move_eigenvalue_informal}] \label{lem:shifted_small_move_eigenvalue_formal}
For 2-layer ReLU activated neural network. Suppose the shifted parameter is $b$ ($b \geq 0$). Let $R_0>0$ be a parameter.  
Suppose 
\begin{align*}
m \ge \Omega(1) \cdot \max\{ b^2R_0^2, n^2 R_0^2 \lambda^{-2}, n\lambda^{-1} \log(n/\rho) \},
\end{align*}
then with prob. $\ge$
$
1-\rho - n^2 \cdot \exp \big( -m\cdot \min\{c'e^{-b^2/2},\frac{R_0}{10\sqrt{m}}\} \big),
$
for {\em every}  $W \in \R^{d \times m}$ satisfying $\max_{r \in [m] }\|w_r - w_r(0)\|_2 \leq R_0/\sqrt{m}$, the following holds
\begin{align*}
\|G_W - G_{W_0}\|_{F} \le \lambda / 2, ~~~~~~ \lambda_{\min}(G_W) \ge \lambda / 2.
\end{align*}
\end{lemma}
\begin{proof}
    We use Lemma \ref{lem:boundary_of_G} by setting $R=R_0/\sqrt{m}$ (that lemma require that $R \leq 1/b$) and letting $W = \left[ \begin{array}{cccc}
        w_1 & w_2 & \cdots & w_m \\  
    \end{array} \right]$.

    Since  $R_0 /\sqrt{m} \le 1/b$, then we have $m \geq R_0^2 b^2$ (this is the corresponding to the first term of $m$ lower bound in lemma statement).
    
    Note $H(W)$ is essentially $G_W$, and $\|w_r(t)-w_r(0)\|_2\le R$ for any $r$, thus by Lemma \ref{lem:boundary_of_G}, we have 
    \begin{itemize}
        \item $\|G_W-G_0\|_F\le n\cdot \min\{ ce^{-b^2/2}, 3R \} = n\cdot \min\{ ce^{-b^2/2}, 3R_0/\sqrt{m} \}$ with prob.
        \begin{align*}
            1-n^2\exp(-m\cdot \min\{c'e^{-b^2/2},R/10\}) = 1-n^2\exp(-m\cdot \min\{c'e^{-b^2/2},\frac{R_0}{10\sqrt{m}}\}),
        \end{align*}
        \item $\lambda_{\min}(G_W)\ge \frac{3}{4}\lambda - n\min\{ ce^{-b^2/2}, 3R \} = \frac{3}{4}\lambda - n\min\{ ce^{-b^2/2}, 3R_0/\sqrt{m} \}$ with prob.
        \begin{align*}
            1-\rho - n^2\exp(-m\cdot\min\{c'e^{-b^2/2},R/10\}) = 1-\rho - n^2\exp(-m\cdot \min\{c'e^{-b^2/2},\frac{R_0}{10\sqrt{m}}\}).
        \end{align*}
    \end{itemize}
    Then it remains to prove 
    \begin{align*}
        n \cdot \min\{ ce^{-b^2/2},3R_0/\sqrt{m} \} \le \frac{\lambda}{2}.
    \end{align*}
    Since $m\ge \Omega(n^2R_0^2\lambda^{-2})$, we have $3nR_0/\sqrt{m} \le \frac{\lambda}{2}$, which finishes the proof.
\end{proof}

\subsection{Proof of Lemma~\ref{lem:small-move_informal}}
\label{sec:app_analysis:lem_small-move_informal}

\begin{lemma}[The shifted $\mathsf{NTK}$ version of Lemma C.4 in \cite{bpsw21}, formal version of Lemma \ref{lem:small-move_informal}] \label{lem:small-move_formal}
	Suppose $R_0 \geq 1$ and $m = \tilde{\Omega}(n^2 R_0^2)$. Then for every $w \in \R^{d \times m}$ satisfying $\max_{r \in [m]} \|w_r - w_r(0)\|_2 \leq R_0/\sqrt{m}$, the following holds
	\begin{itemize}
		\item $\|W - W_0\| = O(R_0)$,
		\item $\|J_{W, x_i} - J_{W_0, x_i}\|_{2} = \tilde{O}  ( {R_0^{1/2}} / {m^{1/4}} )$ and $\|J_{W} - J_{W_0}\|_{F} = \tilde{O} ( {n^{1/2} R_0^{1/2}} / {m^{1/4}} )$,
		\item $\|J_{W}\|_{F} = O(\sqrt{n})$,
	\end{itemize}
	with prob. $\ge 1 - \rho$. The randomness comes from the initialization of $W_0$.
\end{lemma}

\begin{proof}
	(1) The first claim follows from 
	\begin{align*}
	\|W - W_0\| \leq & ~ \|W - W_{0}\|_{F}  \\
	= & ~ \Big( \sum_{r=1}^{m} \|w_r - w_r(0)\|_2^2 \Big)^{1/2} \\
	\leq & ~ \sqrt{m}\cdot R_0/\sqrt{m} \\
	= & ~ R_0.
	\end{align*}
	where the first step comes from $\| \cdot \| \leq \| \cdot \|_F$,  the second step comes from definition of Frobenius norm,  the third step comes from $\|w_r-w_r(0)\|_2\le R_0/\sqrt{m}$, and the last step comes from canceling $\sqrt{m}$. 
	
	(2) For the second claim, we have for any $i \in [n]$
	\begin{align}
	\|J_{W, x_i} - J_{W_0, x_i}\|^2 = & ~\frac{1}{m}\sum_{r=1}^{m}a_r^2\cdot \|x_r\|_2^2\cdot |\textbf{1}_{\langle w_r, x_i \rangle\geq b} - \textbf{1}_{\langle w_r(0), x_i \rangle \geq b}|^2 \notag \\
	= & ~ \frac{1}{m}\sum_{r=1}^{m}|\textbf{1}_{\langle w_r, x_i \rangle\geq b} - \textbf{1}_{\langle w_r(0), x_i \rangle \geq b}|.\label{eq:standard1}
	\end{align}
	The second equality follows from $a_r \in \{-1, 1\}$, $\|x_i\|_2 =1$ and
	\begin{align}
	\label{eq:standard2}
	s_{i, r} := |\textbf{1}_{\langle w_r, x_i \rangle\geq b} - \textbf{1}_{\langle w_r(0), x_i \rangle \geq b}| \in \{0, 1\}.
	\end{align}
	We define the event $A_{i, r}$ as 
	\begin{align*}
	    A_{i, r} = \left\{\exists \tilde{w} ~:~ \|\tilde{w} - w_r(0)\| \leq R_0/\sqrt{m}, ~~~ \textbf{1}_{\langle \tilde{w}, x_i \rangle\geq b} \neq \textbf{1}_{\langle w_r(0), x_i \rangle \geq b} \right\}.
	\end{align*}
	It is not hard to see $A_{i, r}$ holds if and only if $\langle w_r(0),x_i \rangle \in [b-R_0/\sqrt{m}, b+R_0/\sqrt{m}]$. Since $w_r(0)$ is sampled from Gaussian ${\cal N}(0,I_d)$ and $\|x_i\|=1$, we have $\langle w_r(0), x_i \rangle$ is sampled from Gaussian ${\cal N}(0,1)$, thus by the anti-concentration of Gaussian (see Lemma~\ref{lem:anti_gaussian}), we have 
	\begin{align*}
	    \E[s_{i, r}] = \Pr[A_{i, r}] &= \Pr[~\langle w_r(0),x_i \rangle \in [b-R_0/\sqrt{m}, b+R_0/\sqrt{m}]~]\\
	    &\le \Pr[~\langle w_r(0),x_i \rangle \in [-R_0/\sqrt{m}, R_0/\sqrt{m}]~]\\
	    &\le \frac{4}{5}R_0/\sqrt{m}.
	\end{align*}
	Thus we have
	\begin{align}
	    \Pr\left[\sum_{i=1}^{m} s_{i, r} \geq  ( t + 4 / 5 ) R_0\sqrt{m} \right] 
	    \leq & ~ \Pr\left[\sum_{i=1}^{m} (s_{i, r} - \E [ s_{i, r} ] ) \geq  t R_0\sqrt{m} \right] \notag \\
	    \leq & ~ 2\exp\left( -\frac{2t^2R_0^2 m}{m} \right) \notag \\
	    = & ~ 2\exp(-t^2R_0^2) \notag \\
	    \leq & ~ 2\exp(-t^2). \label{eq:standard3}
	\end{align}
	holds for any $t > 0$.
	The second inequality is due to the Hoeffding bound (see Lemma~\ref{lem:hoeffding}), the last inequality is because $R_0 > 1$. Taking $t = 2\log(n / \rho)$ and using union bound over $i$, with prob. $\ge 1 - \rho$,
	\begin{align*}
	\|J_{W, x_i} - J_{W_0, x_i}\|_{2}^2 = \frac{1}{m}\sum_{r=1}^{m}s_{i, r} \leq \frac{1}{m} \cdot 2\log (n/\rho) R_0\sqrt{m} = \tilde{O}(R_0/\sqrt{m})
 	\end{align*}
 	holds for all $i \in [n]$. The first equality comes from Eq.~\eqref{eq:standard1} and Eq.~\eqref{eq:standard2}, the second inequality comes from Eq.~\eqref{eq:standard3}. Thus we conclude with 
 	\begin{align}\label{eq:bound_diff_J_W_and_J_W_0}
 	\|J_{W, x_i} - J_{W_0, x_i}\|_{2} = \tilde{O} ( R_0^{1/2} / m^{1/4} ) \text{~~~and~~~}\|J_{W} - J_{W_0}\|_{F} = \tilde{O} ( n^{1/2} R_0^{1/2}  / m^{1/4} ).
 	\end{align}
 	(3) The thrid claim follows from
 	\begin{align*}
 	\|J_{W}\|_{F} 
 	\leq & ~ \|J_{W_0}\|_{F} + \|J_{W} - J_{W_0}\|_{F} \\
 	\leq & ~ O(\sqrt{n}) + \|J_{W} - J_{W_0}\|_{F} \\
 	\leq & ~ O(\sqrt{n}) + \tilde{O} ( n^{1/2}R_0^{1/2} / m^{1/4} ) \\
 	= & ~ O(\sqrt{n}).
 	\end{align*}
 	where the 1st step is due to triangle inequality, the 2nd step is due to the third claim in Lemma \ref{lem:shifted_initialization_informal}, the 3rd step is due to Eq.~\eqref{eq:bound_diff_J_W_and_J_W_0}, and the last step is due to $m = \tilde{\Omega}(R_0^2n^2)$.
	
\end{proof}

\section{Induction}\label{sec:induction}

Section \ref{sec:correctness} has defined the induction hypothesis (see Definition \ref{def:correctness_hypothesis}) and given a lemma (see Lemma \ref{lem:proof_hypothesis}) to prove that induction hypothesis holds for all time with high probability, but left its proof to this section. Here, we present and prove the following Lemma \ref{lem:proof_hypothesis_formal}, the formal version of Lemma \ref{lem:proof_hypothesis}, and then the crucial Theorem \ref{thm:correctness} holds straightforwardly. We divided the proof of each part of the lemma in Section~\ref{sec:part_I} and Section~\ref{sec:part_II}, and combine them in Section~\ref{sec:lem_combine}.

\begin{lemma}[Formal version of Lemma \ref{lem:proof_hypothesis}] \label{lem:proof_hypothesis_formal}
    Define $R_0 \approx n/\lambda$. With probability at least $1-\frac{5}{2}\rho - n^2 \cdot \exp \big( -m\cdot \min\{c'e^{-b^2/2},\frac{R_0}{10\sqrt{m}}\} \big)$ of the initial weights $W_0$, for every $t>0$, if  
    \begin{itemize}
        \item  $\|f_{t} - y\|_2 \leq \frac{1}{2} \|f_{t-1} - y\|_2$
	    \item $\max_{r \in [m]} \| w_r(t) - w_r(0) \|_2 \leq R_0 / \sqrt{m}$
    \end{itemize}
    then
    \begin{itemize}
        \item  $\|f_{t+1} - y\|_2 \leq \frac{1}{2} \|f_{t} - y\|_2$
        \item $\max_{r\in [m]}\| w_r(t+1) - w_r(0) \|_2 \leq R_0 / \sqrt{m}$
    \end{itemize}
    also holds. 
\end{lemma}

\subsection{Proof of Lemma \ref{lem:proof_hypothesis_formal}: the first lemma} \label{sec:part_I}
As stated in the previous subsection, we use induction. Here we need to break the induction step (Lemma~\ref{lem:proof_hypothesis_formal}) into two separate steps, Lemma~\ref{lem:induction_part1} and Lemma~\ref{lem:induction_part2}.
Each separated induction step corresponds to prove one part in the  Lemma~\ref{lem:proof_hypothesis_formal}.
We first prove the first part of Lemma~\ref{lem:proof_hypothesis_formal}.

\begin{lemma}[Part 1 of Lemma~\ref{lem:proof_hypothesis_formal}]\label{lem:induction_part1}
Suppose initial weights $W_0$ satisfies the restriction of Lemma \ref{lem:shifted_initialization_informal}, \ref{lem:shifted_small_move_eigenvalue_informal} and \ref{lem:small-move_informal}, then for any fixed $t$, if 
\begin{itemize}
    \item  $\|f_{t} - y\|_2 \leq \frac{1}{2} \|f_{t-1} - y\|_2$ holds
	\item $\max_{r \in [m]}\| w_r(t) - w_r(0) \|_2 \leq R_0 / \sqrt{m}$ holds
\end{itemize}
Then we have
\begin{itemize}
    \item  $\|f_{t+1} - y\|_2 \leq \frac{1}{2} \|f_{t} - y\|_2$ holds.
\end{itemize}
\end{lemma}

This proof is similar to \cite{bpsw21}, for the completeness, we still provide the details here.

\begin{proof}
    We prove the first claim holds for time $t+1$. Define
	\begin{align*}
	J_{t, t+1} = \int_{0}^{1} J \Big( (1 - s) W_t + s W_{t+1} \Big) \d s,
	\end{align*}
	and denote $g^{\star} = (J_tJ_t^{\top})^{-1}(f_t - y)$ to be the optimal solution to Eq.~\eqref{eq:regression}, then we have
	\begin{align}
	&~\|f_{t+1} - y\|_2 \notag\\
	= &~ \|f_{t}  - y + (f_{t + 1}  - f_{t})\|_2\notag\\
	= &~ \|f_{t}  - y +J_{t, t+1}(W_{t+1} - W_{t})\|_2\notag\\
	= &~ \|f_{t}  - y -J_{t, t+1}J_t^{\top} g_t\|_2\notag\\
	= &~ \|f_{t}  - y -J_{t}J_t^{\top} g_t + J_{t}J_t^{\top} g_t  -J_{t, t+1}J_t^{\top} g_t\|_2 \notag\\
	\leq &~ \|f_{t}  - y - J_t J_t^{\top} g_t\|_2 +   \|(J_t-J_{t, t+1})J_t^{\top} g_t\|_2\notag\\
	\leq &~ \|f_{t}  - y - J_t J_t^{\top} g_t\|_2 +\|(J_t-J_{t, t+1})J_t^{\top} g^{\star}\|_2 +   \|(J_t-J_{t, t+1})J_t^{\top} (g_t - g^{\star})\|_2 \label{eq:second-order3},
	\end{align}
	where the 2nd step is from the definiton of $J_{t, t+1}$ and simple calculus, the 3rd step is from the updating rule of the algorithm, the 5th step is due to triangle inequality, and the sixth step is because triangle inequality.
	
	For the first quantity in Eq.~\eqref{eq:second-order3}, we have
	\begin{align}
	\label{eq:second-order5}
	\|J_t J_t^{\top}g_t - (f_t - y)\|_2 \leq \frac{1}{6}\|f_t - y\|_2,
	\end{align}
	since $g_t$ is an $\epsilon_0 (\epsilon_0 \leq \frac{1}{6})$ approximate solution to regression problem~\eqref{eq:regression}.
	
	For the second quantity in Eq.~\eqref{eq:regression}, we have
	\begin{align}
	\|(J_t-J_{t, t+1})J_t^{\top} g^{\star}\|_2 
	\leq & ~ \|(J_t-J_{t, t+1})\| \cdot \|J_t^{\top} g^{\star}\|_2\notag\\
	 = & ~ \|(J_t-J_{t, t+1})\| \cdot \|J_t^{\top} (J_t J_t^{\top})^{-1} (f_t - y) \|_2\notag\\
	 \leq & ~ \|(J_t-J_{t, t+1})\| \cdot \|J_t^{\top} (J_t J_t^{\top})^{-1} \| \cdot \| (f_t - y)\|_2 \label{eq:second-order9}
	\end{align}
	where the 1st step is due to matrix spectral norm, the 2nd step is because the definition of $g^*$, and the 3rd step relies on matrix spectral norm.

	We bound these term separately. First,
		\begin{align}
		\|J_t - J_{t, t+1}\|
		\leq & ~ \int_{0}^{1}\| J((1 - s)W_t + sW_{t+1}) - J(W_t) \|\d s \notag\\
		\leq &~ \int_{0}^{1}\left( \| J((1 - s)W_t + sW_{t+1}) - J(W_0)\| + \|J(W_0) - J(W_t) \| \right) \d s \notag\\
		\leq &~ \tilde{O} ( R_0^{1/2} n^{1/2} / m^{1/4} ),
		\label{eq:second-order6}
		\end{align}
	
		where the 1st step comes from simple calculus, the 2nd step comes from triangle inequality, and the 3rd step comes from the second claim in Lemma~\ref{lem:small-move_informal} and the fact that
		\begin{align*}
		\|(1-s)w_r(t) + sw_r(t+1) - w_0\|_2 \leq &~ (1 - s)\|w_r(t) - w_r(0)\|_2 + s\|w_{r}(t + 1) - w_r(0)\|_2\\
		\leq &~ R_0 /\sqrt{m}.
		\end{align*}
	Then, we have
	\begin{align}
		\|J_t^{\top} (J_t J_t^{\top})^{-1}\| = \frac{1}{\sigma_{\min}(J_t^{\top})} \leq \sqrt{ 2 / \lambda } \label{eq:second-order10}
	\end{align}
	where the 2nd step comes from $\sigma_{\min}(J_t) = \sqrt{\lambda_{\min}(J_t^{\top}J_t)} \geq \sqrt{ \lambda / 2 }$ (see Lemma~\ref{lem:shifted_small_move_eigenvalue_informal}). 

	Combining Eq.~\eqref{eq:second-order9}, \eqref{eq:second-order6} and \eqref{eq:second-order10}, we have
	\begin{align}
	\label{eq:second-order11}
	\|(J_t-J_{t, t+1})J_t^{\top} g^{\star}\|_2  &\leq \tilde{O} ( {R_0^{1/2}\lambda^{-1/2}n^{1/2}} / {m^{1/4} } )\|f_t - y\|_2 \notag \\
	&= \tilde{O} ( {\lambda^{-1}nm^{-1/4} } )\|f_t - y\|_2 \notag \\
	&\leq  \|f_t - y\| /6,
	\end{align}
	since $m = \tilde{\Omega}(\lambda^{-4} n^4)$.
	
	Let us consider the third term in Eq.~\eqref{eq:second-order3},
	\begin{align}
	\label{eq:second-order4}
	\|(J_t-J_{t, t+1})J_t^{\top} (g_t - g^{\star})\|_2  \leq\|J_t - J_{t, t+1}\| \cdot \|J_t^{\top}\| \cdot \|g_t - g^{\star}\|_2
	\end{align}
	by matrix norm. Moreover, one has
	\begin{align}\label{eq:second-order2}
	\frac{\lambda}{2} \|g_t - g^{\star}\|_2 
	\leq & ~ \lambda_{\min}(J_t J_t^{\top}) \|g_t - g^{\star}\|_2 \notag \\
	\leq & ~ \|J_t J_t^{\top}g_t - J_t J^{\top}_tg^{\star}\|_2 \notag \\
	= & ~ \|J_t J_t^{\top}g_t - (f_t - y)\|_2 \notag \\
	\leq & ~ \sqrt{ \lambda / n } \cdot \|f_t - y\|_2,
	\end{align}
	where 1st step comes from $\lambda_{\min}(J_tJ_t^{\top}) = \lambda_{\min}(G_t) \geq \lambda / 2$ (see Lemma~\ref{lem:shifted_small_move_eigenvalue_informal}), the 2nd step is because simple linear algebra, the 3rd step is because the definition of $g^*$, and the last step is because $g_t$ is an $\epsilon_0$-approximate solution to $\min_{g_t}\|J_tJ_t^{\top}g_t - (f_t-y)\|$ and $\epsilon_0\le \sqrt{\lambda/n}$. 
	
	 Consequently, we have
	\begin{align}\label{eq:second-order8}
	\|(J_t-J_{t, t+1})J_t^{\top} (g_t - g^{\star})\|_2  
	\leq & ~ \|J_t - J_{t, t+1}\| \cdot \|J_t^{\top}\| \cdot \| g_t - g^{\star} \|_2 \notag \\
	\leq & ~ \tilde{O} (  R_0^{1/2} n^{1/2} m^{-1/4} ) \cdot \sqrt{n} \cdot \frac{2}{\sqrt{n\lambda}} \cdot \| f_t - y \|_2 \notag \\
	= & ~ \tilde{O} ( n \lambda^{-1} m^{-1/4} ) \cdot \|f_t - y\|_2 \notag \\
	\leq & ~ \frac{1}{6} \|f_t - y\|_2,
	\end{align}
	where the 1st step is because of matrix spectral norm, the 2nd step comes from Eq.~\eqref{eq:second-order6}, \eqref{eq:second-order2} and the fact that $\|J_{t}\| \leq O(\sqrt{n})$ (see Lemma~\ref{lem:small-move_informal}), and the last step comes from the $m = \Omega(n^4\lambda^{-4})$.  
	Combining Eq.~\eqref{eq:second-order3}, \eqref{eq:second-order5}, \eqref{eq:second-order11}, and \eqref{eq:second-order8}, we have proved the first claim, i.e.,
	\begin{align}
	\label{eq:second-order12}
	\|f_{t + 1} - y\|_2 \leq \frac{1}{2}\|f_t - y\|_2.
	\end{align}
	Thus, we complete the proof. 
\end{proof}

\subsection{Proof of Lemma \ref{lem:proof_hypothesis_formal}: the second lemma} \label{sec:part_II}
We now move to the second part for Lemma~\ref{lem:proof_hypothesis_formal}. We show it in Lemma~\ref{lem:induction_part2}.

\begin{lemma}[Part 2 of Lemma~\ref{lem:proof_hypothesis_formal}]\label{lem:induction_part2}
Suppose initial weights $W_0$ satisfies the restriction of Lemma \ref{lem:shifted_initialization_informal}, \ref{lem:shifted_small_move_eigenvalue_informal} and \ref{lem:small-move_informal},
then for any fixed $t$, if 
\begin{itemize}
    \item  $\|f_{t} - y\|_2 \leq \frac{1}{2} \|f_{t-1} - y\|_2$ holds
	\item $\max_{r \in [m]}\| w_r(t) - w_r(0) \|_2 \leq R_0 / \sqrt{m}$ holds
\end{itemize}
Then we have
\begin{itemize}
	\item $\max_{r \in [m]}\| w_r(t+1) - w_r(0) \|_2 \leq R_0 / \sqrt{m}$ holds
\end{itemize}
\end{lemma}
This proof is similar to \cite{bpsw21}, for the completeness, we still provide the details here.

\begin{proof}
	First, we have
	 	 \begin{align}\label{eq:second-order7}
	 	 \|g_t\|_2 
	 	 \leq & ~ \|g^{\star}\|_2 + \|g_t - g^{\star}\|_2 \notag \\
	 	 \leq & ~ \|(J_tJ_t^{\top})^{-1}(f_t - y)\|_2 + \|g_t - g^{\star}\|_2 \notag \\
	 	 \leq & ~ \|(J_tJ_t^{\top})^{-1}\| \cdot \|(f_t - y)\|_2 + \|g_t - g^{\star}\|_2 \notag \\
	 	 \leq & ~ \frac{2}{\lambda}\cdot \|f_t - y\|_2 + \frac{2}{\sqrt{n\lambda}} \cdot \|f_t - y\|_2 \notag \\
	 	 \lesssim & ~ \frac{1}{\lambda} \cdot \|f_t - y\|_2,
	 	 \end{align}
	 where the 1st step relies on triangle inequality, the 2nd step replies on the definition of $g^*$, the 3rd step uses matrix norm, the 4th step comes from Eq.~\eqref{eq:second-order2} and the last step uses the obvious fact that $1/\sqrt{n\lambda} \leq 1/\lambda$.
		 
    Hence, for any $0 \leq k\leq t$ and $r\in [m]$, if we use $g_{k, i}$ to denote the $i^{th}$ indice of $g_k$, then we have
	 
	\begin{align}
		 \|w_r(k+1) - w_r(k)\|_2 = &~ \|(J_k^{\top}g_k)_r\|_2 \notag\\
		 = &~ \left\|\sum_{i=1}^{n}\frac{1}{\sqrt{m}}a_r x_i^{\top}{\bf 1}_{\langle w_r(t), x_i\rangle\geq b}g_{k, i}\right\|_2 \notag\\
		 \leq &~ \frac{1}{\sqrt{m}}\sum_{i=1}^{n}|g_{k, i}| \notag\\
		 \leq &~ \frac{\sqrt{n}}{\sqrt{m}}\|g_k\|_2 \notag\\
		 \lesssim &~ \frac{\sqrt{n}}{\sqrt{m}} \cdot \frac{1}{2^{k}\lambda}\|f_0 - y\|_2 \notag\\
		 \lesssim &~ \frac{n}{\sqrt{m}\lambda} \cdot \frac{1}{2^k} \label{eq:weight_change},
	 \end{align}
	 where the 1st step is because of the updating rule, the 2nd step is because of the definition of $J_k$, the 3rd step is because of triangle inequalities and the fact that $a_r = \pm 1$, $\|x_r\|_2 = 1$, the 4th step comes is because of Cauchy-Schwartz inequality, the 5th step is because of Eq.~\eqref{eq:second-order12} and Eq.~\eqref{eq:second-order7}, and the last step is because of the fact that $\|f_0 - y\|_2 \leq O(\sqrt{n})$ (see Lemma~\ref{lem:shifted_initialization_informal}, $f_0(x_i)=O(1)$ for any $i\in [n]$, thus $\|f_0-y\|_2=\sqrt{\sum_{i=1}^n(f(x_i)-y_i)^2}=O(\sqrt{n}$).
	 Consequently, we have
	 \begin{align*}
	 \|w_r(t+1) - w_r(0)\|_2 \leq &~ \sum_{k=0}^{t}\|w_r(k+1) - w_r(k)\|_2 \lesssim \sum_{k=0}^{t} \frac{n}{\sqrt{m}\lambda} \cdot \frac{1}{2^k} \lesssim \frac{R_0}{\sqrt{m}},
	 \end{align*}
	 where the 1st step is because of triangle inequality, the 2nd step is because of Eq. \eqref{eq:weight_change}, and the last step is because of simple summation.
	 
	 Thus we also finish the proof of the second claim.
\end{proof}

\subsection{Proof of Lemma \ref{lem:proof_hypothesis_formal}: combination}
\label{sec:lem_combine}
We use Lemma \ref{lem:induction_part1} and Lemma \ref{lem:induction_part2} to prove Lemma \ref{lem:proof_hypothesis_formal}.

\begin{proof}
Since the probability of initial weight $W_0$ satisfies the restriction of Lemma \ref{lem:shifted_initialization_informal}, Lemma \ref{lem:shifted_small_move_eigenvalue_informal} and Lemma \ref{lem:small-move_informal} is $1-\rho/2$, $1-\rho - n^2 \cdot \exp \big( -m\cdot \min\{c'e^{-b^2/2},\frac{R_0}{10\sqrt{m}}\} \big)$, $1-\rho$ respectively, by union bound, the probability of they all happen is at least 
\begin{align*}
    1-\frac{5}{2}\rho - n^2 \cdot \exp \big( -m\cdot \min\{c'e^{-b^2/2},\frac{R_0}{10\sqrt{m}}\} \big)
\end{align*}

In this case, for any fixed $t$, combining Lemma \ref{lem:induction_part1} and Lemma \ref{lem:induction_part2}, if 
\begin{itemize}
    \item  $\|f_{t} - y\|_2 \leq \frac{1}{2} \|f_{t-1} - y\|_2$ holds,
	\item $\max_{r \in [m]} \| w_r(t) - w_r(0) \|_2 \leq R_0 / \sqrt{m}$ holds
\end{itemize}
then we have
\begin{itemize}
    \item  $\|f_{t+1} - y\|_2 \leq \frac{1}{2} \|f_{t} - y\|_2$ holds.
    \item $\max_{r \in [m]}\| w_r(t+1) - w_r(0) \|_2 \leq R_0 / \sqrt{m}$ holds
\end{itemize}

Thus by induction, with prob. $\ge 1-\frac{5}{2}\rho - n^2 \cdot \exp \big( -m\cdot \min\{c'e^{-b^2/2},\frac{R_0}{10\sqrt{m}}\} \big)$, 
\begin{align*}
    \|f_t-y\|_2 \le \frac{1}{2}\|f_{t-1}-y\|_2
\end{align*}
holds for all $t$, hence finished the proof of Lemma \ref{lem:proof_hypothesis_formal}.
\end{proof}

\subsection{Number of Iterations for Iterative Regression}
\begin{lemma}[]\label{lem:iterative_regression_number_of_iterations}
The iterative regression in our fast training algorithm requires $O(\log(n/\lambda))$ iterations.
\end{lemma}
\begin{proof}
By Lemma \ref{lem:shifted_initialization_formal}, $\|J_tJ_t^\top\| = \|G_t\| = O(n)$ and $\lambda_{\min}(J_tJ_t^\top) = \lambda_{\min}(G_t) \ge O(\lambda)$. Let $\epsilon_{\mathrm{reg}}$ be chosen as  Algorithm \ref{alg:ours_formal}.

Thus by Corollary \ref{cor:iterative_regression}, the number of iterations needed by the iterative regression is 
\begin{align*}
   O(\log (\kappa(J_t^\top)/\epsilon_{\mathrm{reg}}) ) = & ~ O( \log (\sqrt{n/\lambda} / \sqrt{\lambda/n}) ) \\
   = & ~ O(\log(n/\lambda)).
\end{align*}
\end{proof}

\section{More Running Time Details}\label{sec:app_time}

Section \ref{sec:running_time} analyzes the running time of our algorithm. It shows that when $m$ is large enough, the running time of $\mathsf{CPI}$ is $o(mnd)+\wt{O}(n^3)$, and with $\mathsf{FMM}$, the $\mathsf{CPI}$ can be reduced to $o(mnd)+\wt{O}(n^\omega)$.

In this section, we give the specific time complexity hidden by $o(mnd)$, and also give the complete algorithm representation of our training algorithm. It will show that when $m$ is large enough, the $\mathsf{CPI}$ is $\wt{O}(m^{1-\alpha}nd)$. 
Similar with Section \ref{sec:running_time}, we first present Theorem \ref{thm:running_time_formal}, the running time result. We then provide three lemmas (Lemma~\ref{lem:sketch_time_formal}, Lemma~\ref{lem:regression_time_formal} and Lemma~\ref{lem:backward_time_formal}) to prove our main theorem.
Our main running time result is the following:

\begin{theorem}[Running time part of Theorem \ref{thm:main}, formal version of Theorem \ref{thm:running_time_informal}] \label{thm:running_time_formal}

The $\mathsf{CPI}$ is $\tilde{O}( m^{1-\alpha}nd + n^3 )$, and the running time for shrinking the training loss to $\epsilon$ is $\tilde{O}( (m^{1-\alpha}nd + n^3) \log(1/\epsilon))$.

Using $\mathsf{FMM}$, the $\mathsf{CPI}$ is $\tilde{O}( m^{1-\alpha}nd + n^{\omega} )$, the running time is $\tilde{O}( (m^{1-\alpha}nd + n^\omega)\log(1/\epsilon))$. Note that $\omega$ is the exponent of matrix multiplication. Currently, $\omega \approx 2.373$.
\end{theorem}

\begin{proof}
Combining Lemma \ref{lem:sketch_time_formal}, Lemma \ref{lem:regression_time_formal} and Lemma \ref{lem:backward_time_formal}, the computation time of each iteration is 
\begin{align*}
      & ~\tilde{O}( n^2m^{0.76}d ) + \tilde{O}( nm^{0.76}d + n^3 ) + O( n^2m^{0.76}(d+\log m) ) \\
    = & ~\tilde{O}( n^2m^{0.76}d + n^3 + n^2m^{0.76}d ) \\
    = & ~\tilde{O}( n^2m^{0.76}d + n^3 ),
\end{align*}
where the first step comes from hiding $\log m$ on $\tilde{O}$, the second step comes from simple merging. 
And if using $\mathsf{FMM}$, similarly the running time is $\tilde{O}( n^2m^{0.76}d + n^\omega )$.

By Theorem \ref{thm:correctness}, we have:
The time to reduce the training loss to $\epsilon$ is $\tilde{O}( (n^2m^{0.76}d + n^3 ) \log(1/\epsilon))$. Taking advantage of $\mathsf{FMM}$, the time is $\tilde{O}( (n^2m^{0.76}d + n^\omega ) \log(1/\epsilon))$.

Further, for example, if $m = n^{c}$ where $c$ is some large constant, then $n^2 m^{0.76} d \leq nm^{1-\alpha}d$ where $\alpha \in [0.1,0.24)$. 
Hence the time of each iteration is $\tilde{O}( m^{1-\alpha}nd + n^3 )$, and the time to reduce the training loss to $\epsilon$ is $\tilde{O}( (m^{1-\alpha}nd + n^3 ) \log(1/\epsilon))$. Taking advantage of $\mathsf{FMM}$, the time is $\tilde{O}( (m^{1-\alpha}nd + n^\omega )\log(1/\epsilon))$. Thus we complete the proof.
\end{proof}

For the rest of this section, we provide detailed analysis for the steps. In Section~\ref{sec:sketch_compute} we analyse the sketch computing step. In Section~\ref{sec_iterative_reg} we analyse the iterative regression step. In Section~\ref{sec:weight_maint} we analyse the implicit weight maintenance step. 

\subsection{Sketch computing} 
\label{sec:sketch_compute}
We delicate to prove the lemma that formally analyzes the running time of the sketch computing process in Algorithm \ref{alg:ours_formal} to show its time complexity.

\begin{lemma}[Sketch computing, formal version of Lemma \ref{lem:sketch_time}]\label{lem:sketch_time_formal}
The sketch computing process of Algorithm \ref{alg:ours_formal} (from line \ref{line:sketch_start} to line \ref{line:sketch_end}) runs in time
$
\wt{O}(m^{0.76}n^2d ).
$
\end{lemma}
\begin{proof}
In the sketch computing process, by Corollary \ref{cor:sparse}, only $O(m^{0.76}d)$ 
entries of each column of $A$ is nonzero, thus calculating each column of $B$ takes $O(m^{0.76}dt)$ 
time, where $t$ is the number of rows of $B$. And according to Lemma \ref{lem:subspace-embedding}, 
\begin{align*}
t= & ~ n \poly( \log(n/\delta_{\sketch}) ) / \epsilon_{\sketch}^2 \\
= & ~ O(n \poly( \log(n/\delta_{\sketch}) )).
\end{align*}
Since 
\begin{align*}
\epsilon_{\sketch}=0.1
\mathrm{~~~and~~~} \delta_{\sketch}=\frac{1}{\poly(n)},
\end{align*}
the whole for-loop runs in time 
$
O(n^2m^{0.76}d\poly(\log(n))).
$
\end{proof}

\subsection{Iterative regression}
\label{sec_iterative_reg}
We delicate to prove a lemma that formally analyzes the running time of the iterative regression process in Algorithm \ref{alg:ours_formal} to show its time complexity.

\begin{lemma}[Iterative regression, formal version of Lemma \ref{lem:regression_time}]\label{lem:regression_time_formal}
The iterative regression of Algorithm \ref{alg:ours_formal} (from line \ref{line:iter_start} to line \ref{line:iter_end}) runs in time
\begin{align*}
\wt{O}( nm^{0.76}d  + n^3).
\end{align*}
Taking advantage of $\mathsf{FMM}$, the running time is 
\begin{align*}
\wt{O}( nm^{0.76}d + n^\omega),
\end{align*}
where $\omega$ is the exponent of matrix multiplication. Currently $\omega \approx 2.373$ \cite{w12}.
\end{lemma}
\begin{proof}
The algorithm calculate $R$ using $QR$ decomposition in line \ref{row:QR_decomposition} (Algorithm~\ref{alg:ours_formal}). This step will take $O(n^3)$ time. Taking advantage of $\mathsf{FMM}$, it will take $O(n^\omega)$ time \cite{w12}.

For the while-loop from line \ref{row:iterative_regression} (Algorithm~\ref{alg:ours_formal}), define $p$ as the number of iterations of the while-loop from line \ref{row:iterative_regression} (Algorithm~\ref{alg:ours_formal}), then
\begin{align*}
p &= O(\log (n/\lambda) ) \\
&= O(\log (\frac{n}{(\exp(-b^2/2)\cdot \frac{\delta}{100n^2})} ) )\\
&= O(\log (n/\delta) + b^2 )\\
&= O(\log (n/\delta) + \log m)\\
&= O(\log (mn/\delta)),
\end{align*}
where the 1st step comes from Lemma \ref{lem:iterative_regression_number_of_iterations}, the 2nd step comes from Theorem \ref{thm:sep}, the 3rd step comes from identical transformation, and the 4th step comes from $b= \Theta( \sqrt{\log m} )$.

And in each iteration, note that $R$ is $n\times n$, $A$ is $md\times n$, $S$ is $t\times md$,
\begin{itemize}
\item we have calculating $v = R^\top A^\top (D \otimes I_d) A R z_t - R^\top y_{\mathrm{reg}}$ takes
\begin{align*}
    O(n^2+m^{0.76}dn+m^{0.76}dn+n^2+n^2)=O(m^{0.76}dn+n^2)
\end{align*} 
time, 
\item and calculating $(R^\top A^\top (D \otimes I_d) A R)^\top v$ takes
\begin{align*}
    O(n^2+m^{0.76}dn+m^{0.76}dn+n^2)=O(m^{0.76}dn+n^2)
\end{align*}
time.
\end{itemize}
Thus each iteration in the while-loop from line \ref{row:iterative_regression} (Algorithm \ref{alg:ours_formal}) takes
$O(m^{0.76}dn+n^2)$ 
time, the total process of the iterative regression takes
$O((m^{0.76}dn+n^2)\log(mn/\delta) + n^3)$ 
time. 

Using $\mathsf{FMM}$, the running time is
$O((m^{0.76}dn+n^2)\log(mn/\delta) + n^\omega)$.

In our regime, $O(\log (n/\delta)) = O(\log m)$ since $m = \poly(n/\delta)$. Thus, we can hide the log factors in $\wt{O}$.
\end{proof}

\subsection{Implicit weight maintenance} 
\label{sec:weight_maint}
We give a lemma that formally analyzes the running time of the implicit weight maintenance process in Algorithm \ref{alg:ours_formal} to show its time complexity.

\begin{lemma}[Implicit weight maintenance, formal version of Lemma \ref{lem:backward_time}] \label{lem:backward_time_formal}

The implicit weight maintenance of Algorithm \ref{alg:ours_formal} (from line \ref{line:maintenance_start} to line \ref{line:maintenance_end}) runs in time 
$
O(n^2m^{0.76}(d+\log m)).
$
\end{lemma}
\begin{proof}
Let us consider every iteration of the for loop starting at line \ref{row:backward_computation} (Algorithm \ref{alg:ours_formal}), since $(J_t)_{r}$ is $d\times n$, computing $W_{t+1}$ takes $O(nd)$ time. And by Lemma \ref{lem:update_time}, updating $W_{t+1}$ takes $O(n(d+\log m))$ time, thus each iteration takes $O(n(d+\log m))$ time. By Theorem \ref{thm:neural_network_sparisity},
$|K|=O(nm^{0.76})$,
thus the whole implicit weight maintenance takes
$O(n^2m^{0.76}(d+\log m))$ 
time.
\end{proof}

\section{Combination} \label{sec:combination}
Theorem \ref{thm:main} shows that as long as the 2-layer neural network is broad enough, then there exists a training algorithm with sublinear running time and large converge probability. Theorem \ref{thm:correctness} gives an analysis about how large $m$ should be, but its result is based on $\lambda$, the minimal eigenvalue of $K$ \footnote{See Section \eqref{eq:kernel} for the definition of $K$.}, which is not straightforward.

In this section, we convert the bound of Theorem \ref{thm:correctness} into a bound only related to batch number $n$, data separability $\delta$ and tolerable probability of failure $\rho$.

\begin{definition}[Two sparsity definitions] \label{def:sparsity}

We define sparsity of the 2-layer neural network to the number of activated neurons.

We define sparsity of a Jacobi matrix of 2-layer neural network as the maximal number of non-zero entries of a row in the Jacobi matrix $J$ ($J\in \mathbb{R}^{n\times md}$) of weights. 
\end{definition}

We present the following theorem.

\begin{theorem} \label{thm:neural_network_sparisity}
For a 2-layer ReLU activated neural network. Suppose $m$ is the number of neurons, $d$ is the dimension of points, $n$ is represent the number of points, $\rho \in (0,1/10)$ is the failure probability, and $\delta$ is the separability of data points. 

For any real number $\ov{\alpha} \in (0,1]$, let $b = \sqrt{0.5 (1-\ov{\alpha}) \log m }$, if
\begin{align*}
    m = \Omega( ( \delta^{-4} n^{12} \log^{4} (n/\rho) )^{1/\ov{\alpha}} ) 
\end{align*}
then the training algorithm in Algorithm \ref{alg:ours_formal} converges with prob. $\ge 1-\frac{5}{2}\rho-n^2 \cdot \exp( -m\cdot \min\{c'e^{-b^2/2},\frac{R}{10\sqrt{m}}\} )$, and the sparsity of the neural network is
\begin{align*}
    O(m^{\frac{3+\ov{\alpha}}{4}})
\end{align*}
with probability $1-n\cdot \exp(-\Omega(m \cdot \exp(-b^2/2)))$.
Especially, for any given parameter $\epsilon_0 \in (0,1/4]$, if we choose $\ov{\alpha} = 0.04$, the sparsity is $O(m^{0.76})$.
\end{theorem}

\begin{proof}
From Theorem~\ref{thm:sep}, 
we know
\begin{align*}
    \lambda \geq \exp(-b^2/2) \cdot \frac{\delta}{100 n^2}.
\end{align*}
Since by Theorem \ref{thm:correctness},  
we need 
\begin{align*}
    m = \Omega(\lambda^{-4} n^4 b^2 \log^2(n/\rho) )
\end{align*}
to make our algorithm converges, we need to choose 
\begin{align*}
    m = & ~  \Omega( (\exp(b^2/2) \cdot {100n^2} \cdot {\delta}^{-1} )^4 \cdot  n^4 b^2 \log^2(n/\rho) ) \\
    = & ~\Omega( \exp(4 \cdot b^2/2) \cdot \delta^{-4} \cdot n^{12} b^2 \log^2(n/\rho) ) \\
    = & ~ \Omega( m^{1-\ov{\alpha}} \cdot \delta^{-4} \cdot n^{12} \cdot (\log m) \cdot \log^2(n/\rho) )
\end{align*}
where the final step is because $b = \sqrt{ 0.5(1-\ov{\alpha}) \log m }$.

Suppose the constant hidden by $\Omega$ is $C$, then the above equation is equivalent to 
\begin{align*}
    m^{\ov{\alpha}} \ge C \cdot \delta^{-4} \cdot n^{12} \cdot (\log m) \cdot \log^2(n/\rho), 
\end{align*}
and since $m=\poly(n)$, $\log m \le \log^2 n$, thus as long as
\begin{align*}
    m \geq  ( C \delta^{-4} n^{12} \log^4 (n/\rho) )^{1/\ov{\alpha}},
\end{align*}
we have $m = \Omega(\lambda^{-4} n^4 b^2 \log^2(n/\rho) )$, then by Theorem \ref{thm:correctness}, our algorithm converges.

Then, according to Lemma \ref{lem:sparse_initial}, the sparsity of this neural network 
is equal to 
\begin{align*}
= & ~ O(m \cdot \exp(-b^2/2)) \\
= & ~ m \cdot m^{-(1-\ov{\alpha})/4} \\
= & ~ m^{\frac{3+\ov{\alpha}}{4}}
\end{align*}
where the second step is because $b = \sqrt{ 0.5(1-\ov{\alpha}) \log m }$, for any $\ov{\alpha} \in (0,1]$.
\end{proof}

\ifdefined\isarxiv

\bibliographystyle{alpha}
\bibliography{ref}
\else

\fi 



\end{document}